%% file: paper.tex
\PassOptionsToPackage{capitalise}{cleveref}
\documentclass[]{gtech}
\usepackage{latexml} % Official PDF/LaTeXML conditional support.

\usepackage{amsmath,amssymb}
\usepackage{multirow,bigdelim}
\usepackage{longtable}
\usepackage{tabularray}
\UseTblrLibrary{booktabs}
\usepackage{wrapfig}
\usepackage[most]{tcolorbox}
\usepackage{xcolor}
\usepackage{url}
\usepackage{colortbl}
\usepackage{caption}
\usepackage{pifont}
\usepackage{booktabs}
\usepackage{makecell}
\usepackage{tabulary}
\usepackage{multicol}
\usepackage{graphicx}
\usepackage{microtype}
\usepackage{enumitem}
\usepackage{placeins}
\usepackage{flafter}
\usepackage{xspace}
\usepackage{tabularx}
\usepackage{array}

\makeatletter
\@ifpackageloaded{natbib}{}{\usepackage[authoryear,round,sort]{natbib}}
\AtBeginDocument{%
  \@ifundefined{NAT@longnamesfalse}{}{\NAT@longnamesfalse}%
  \def\NAT@sort{\@ne}%
  \setcitestyle{authoryear,round,semicolon,aysep={,},yysep={,}}%
  \let\cite\citep
  \let\citet\citep
  
}
\makeatother

\ifdefined\DeclareUnicodeCharacter
  \DeclareUnicodeCharacter{FF0C}{,}
\fi

\renewcommand{\title}[1]{%
  \newcommand{\titlelist}{{\huge\fontfamily{optimistic}\fontseries{m}\selectfont #1}}}

\newcommand{\method}{SleuthVLM\xspace}
\DeclareMathOperator*{\argmax}{arg\,max}
\newcommand{\cmd}[1]{\texttt{#1}}                 % sandbox file, tool or command, e.g. \cmd{vsheet}
\newcommand{\Budget}{\mathcal{B}}

\definecolor{oursbg}{HTML}{EAF2FB}                % background of our rows
\providecolor{ForestGreen}{cmyk}{0.91,0,0.88,0.12}
\newcommand{\ours}{\rowcolor{oursbg}}
\newcommand{\best}[1]{\textbf{#1}}
\newcommand{\second}[1]{\underline{#1}}
\newcommand{\na}{\textcolor{gray}{--}}            % not reported
\newcommand{\group}[1]{\multicolumn{7}{l}{\textit{#1}}}
\newcommand{\dopen}[1]{\,{\scriptsize\textcolor{ForestGreen}{(#1)}}}       % gap to the best open-source omni model
\newcommand{\dclosed}[1]{\,{\scriptsize\textcolor{red!80!black}{(#1)}}}    % gap to the best closed-source omni model

\title{\textcolor[HTML]{0369FF}{SleuthVLM}: Do We Really Need Audio-in-Video for Omni Video Understanding?}

\author[1,2]{Peiran~Wu}
\author[2,3]{Fangfei~Gou}
\author[2]{Zhaoxi~Mu}
\author[2]{Changhao~Zhang}
\author[3]{Yuntao~Wang}
\author[3]{Yuanchun~Shi}
\author[2]{Weiqiang~Wang}
\author[2,\dagger]{Jian~Liu}
\author[1]{Junxiao~Shen}

\affiliation[1]{University of Bristol}
\affiliation[2]{Ant Group}
\affiliation[3]{Tsinghua University}

\contribution[\dagger]{Corresponding author}

\abstract{%
Omni models answer questions about videos by jointly processing sampled frames and audio in one turn, while VLMs have become strong agents that can search a video over many turns.
We ask how far a model can go without hearing the audio, and present \method, a simple multi-turn harness built on such a VLM that never hears the audio directly: speech is transcribed into text, and the VLM sleuths the video without a predefined workflow.
On four omni benchmarks with videos up to 90 minutes, \method achieves the best average accuracy: \textbf{75.0\%}, 2.6 points above Gemini 3.8 Flash, and it tops the Video-MME-v2 leaderboard with \textbf{77.8\%}; with Qwen3.8-27B, it is comparable to open-source omni models on short videos and better on long ones.
We draw three insights: searching outperforms passive observation; agentic ability matters as much as direct audio access; and small agentic models are highly competitive. So we provide a simple but effective starting point for omni video understanding.
}
\definecolor{ProjectLabelColor}{HTML}{000000}
\definecolor{ProjectLinkColor}{HTML}{0064E0}

\newcommand{\ProjectResourceLabel}[1]{%
  {\fontfamily{ptm}\fontseries{b}\selectfont
   \textcolor{ProjectLabelColor}{#1}}%
}
\newcommand{\ProjectResourceURL}[2]{%
  \href{#1}{\textcolor{ProjectLinkColor}{\textnormal{#2}}}%
}
\newcommand{\ProjectLinks}{%
  \begingroup
  \normalfont\fontsize{9.5}{12.5}\selectfont
  \setlength{\parindent}{0pt}%
  \setlength{\parskip}{0pt}%
  \renewcommand{\arraystretch}{1.08}%
  \noindent
  \begin{tabular}{@{}l@{\hspace{0.9em}}l@{}}
    \ProjectResourceLabel{Project:} &
    \ProjectResourceURL{https://wpr001.github.io/sleuthVLM/}{https://wpr001.github.io/sleuthVLM/}\\[4pt]
    \ProjectResourceLabel{Code:} &
    \ProjectResourceURL{https://github.com/WPR001/OpenOmniHarness}{https://github.com/WPR001/OpenOmniHarness}%
  \end{tabular}\par
  \endgroup
}

\newlength{\ProjectLogoRaise}
\newcommand{\ProjectFooter}{%
  \noindent
  \begin{minipage}[c]{\dimexpr\linewidth-2.6cm\relax}
    \ProjectLinks
  \end{minipage}%
  \hfill
  \begin{minipage}[c]{2cm}
    \centering
    \raisebox{\ProjectLogoRaise}[\height][\depth]{%
      \includegraphics[width=\linewidth]{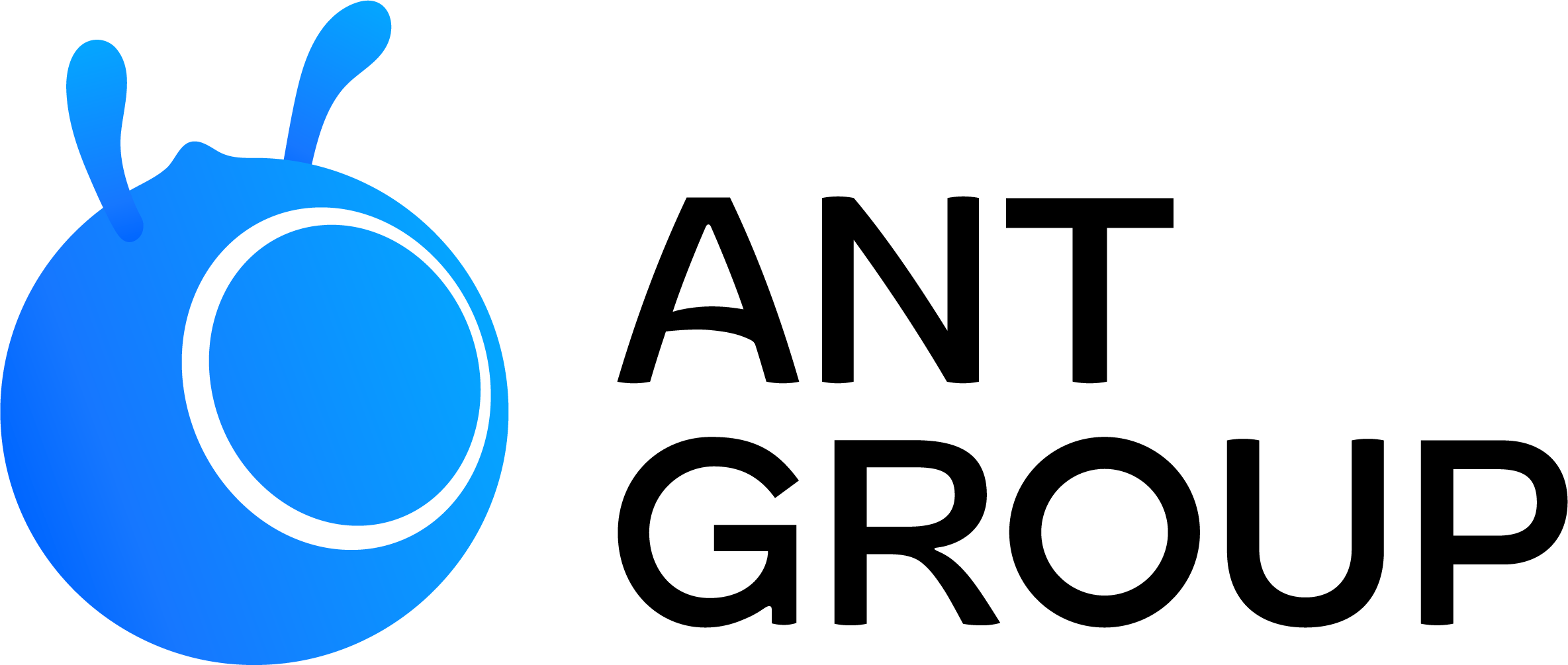}%
    }
  \end{minipage}\par
}

\date{}
\ifdefined\gtechdatalist
  \renewcommand{\gtechdatalist}{}%
\fi
\iflatexml
\else
  \patchcmd{\mymaketitle}
    {\begin{tcolorbox}}
    {\begin{tcolorbox}[overlay={}]}
    {}%
    {\PackageError{sleuthvlm}{Could not disable the fixed title logo}%
      {Adapt the title-overlay patch to the current gtech.cls before compiling.}}%
  \renewcommand{\gtechdatalist}{\ProjectFooter}%
\fi

\begin{document}
\maketitle
\iflatexml
  \begin{tcolorbox}[
    colback=cyan!7!white,
    colframe=cyan!7!white,
    boxrule=0pt,
    arc=10pt,
    left=12pt,right=12pt,top=8pt,bottom=8pt
  ]
    \ProjectFooter
  \end{tcolorbox}
\fi

\begin{figure}[!htbp]
  \centering
  \includegraphics[width=\linewidth]{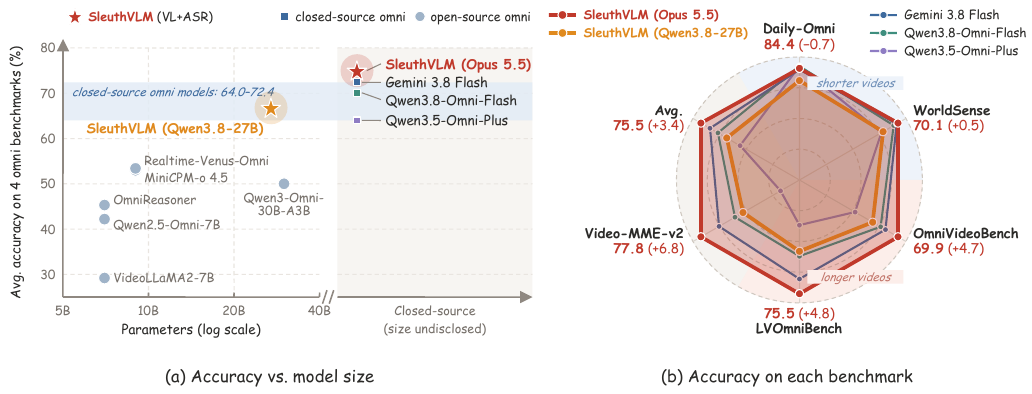}
  \caption{\method never hears the audio directly and sets a new state of the art.
  (a)~Average accuracy on four omni benchmarks against model size; closed-source models are on the right, and the blue band spans the closed-source omni models.
  (b)~Accuracy on the four omni benchmarks, Video-MME-v2 and their average; numbers give \method (Opus 5.5) and, in parentheses, its difference from the best closed-source omni model.}
  \label{fig:teaser}
\end{figure}
\FloatBarrier

\input{sec/1_intro}
\input{sec/2_related}
\input{sec/3_method}
\input{sec/4_experiments}

\input{sec/5_cases}
\input{sec/6_conclusion}

\clearpage
\bibliographystyle{plainnat}
\bibliography{main}

\newpage
\beginappendix
\input{sec/appendix}

\end{document}

%% file: sec/1_intro.tex
\section{Introduction}
\label{sec:intro}

Many questions about videos require understanding both visual and audio content. Omni models address this task by combining a vision-language model (VLM) with an audio encoder and training on paired omni data, enabling them to answer questions from sampled frames and the audio track in one turn.
Closed-source models such as Gemini 3.8 Flash, Qwen3.8-Omni-Flash~\citep{team2026qwen3} and many open-source models follow this recipe, and the field improves them mainly by making model size bigger and using more data~\citep{tang2025video,zhang2024llava,wang2025internvl3,wu2025ugc,cui2026minicpm}.

We revisit two design choices in this recipe.
First, it requires the model to hear the audio directly~\citep{xu2025qwen3,cui2026minicpm,tang2025video}.
In practice, most of the audio that questions depend on is speech (75.2\% of the questions of LVOmniBench~\citep{tao2026lvomnibench}), and speech is easily turned into text with time stamps, and other sounds into captions.
Teaching a model to hear makes every new skill depend on costly joint training~\citep{chowdhury2024meerkat,jiang2025specific}, while strong VLMs already bring the reasoning and tool use that search needs.
Second, it answers in a single turn.
A fixed sampling rate covers the whole video thinly, while the answer often sits in a few seconds~\citep{bai2023qwen,liu2023visual,wu2026st}; in a long video those seconds are easily missed, but this kind of model cannot go back~\citep{wu2024longvideobench,chen2025longvila,fu2025video,yang2025egolife}.

Many works in video understanding have recognized this problem~\citep{zhang2026deep,fan2024videoagent,ren2026videorag,xing2026native,tao2025active}.
The Qwen team runs Qwen3.8-Omni-Flash and Gemini 3.8 Flash as agents in the Qwen Code harness and gains up to 10.3 points on long videos, and Video-Index~\citep{song2026video} finds that agent tools add about 20 points over a fixed input.
We agree: a model should sleuth a video for the evidence it needs rather than take in everything at once.
This points to a promising direction for omni models as well: models such as Qwen3-Omni, MiniCPM-o and video-SALMONN stand to gain much by moving from one turn to agentic understanding.

Building on the agentic ability of frontier VLMs, we propose \method, which answers audio-visual questions with a VLM that never hears the audio directly (\cref{fig:method}).
The audio is transcribed once with word-level timestamps, and the VLM uses a small set of tools to search the video and transcript.
We specify only its accessible resources, a brief agentic guide, and a budget; the VLM decides where to look, how closely to inspect, and when to stop, using the transcript as an index to locate relevant moments in long videos.

\method establishes a new state of the art (\cref{fig:teaser}), achieving an average accuracy of 75.0\% across four omni benchmarks, 2.6 points above the strongest closed-source omni model, Gemini 3.8 Flash. It leads every metric on the Video-MME-v2 leaderboard, reaching 77.8\% accuracy, 6.8 points above Gemini 3.8 Flash. With Qwen3.8-27B, it is comparable to open-source omni models on short videos and better on long ones.
Our experiments reveal three insights:
\begin{itemize}
  \item \textbf{Searching outperforms passive observation.}
  Agentic inference improves accuracy by up to 10.3 points, and our simple harness outperforms agents equipped with specialized tools.
  \item \textbf{Agentic ability matters as much as direct audio access.}
  Opus 5.5, reading transcripts alone, is the strongest reasoning model in our harness, ahead of omni models that can also listen, and agentic ability lifts omni models as well.
  \item \textbf{Small agentic models are highly competitive.}
  An open 27B VLM running on a single GPU matches the performance of closed-source omni models.
\end{itemize}
Taken together, \textbf{\method provides a simple yet effective starting point for agentic omni video understanding.}

%% file: sec/2_related.tex
\section{Related Work}
\label{sec:Related}

\noindent\textbf{Omnimodal Large Language Models.}
Omnimodal large language models aim to understand video, audio, and text within a single model~\citep{xu2025qwen3,cui2026minicpm,tang2025video,xie2024mini}.
They typically extend a vision-language model (VLM) with an audio encoder and train on large amounts of paired audio-visual data, enabling them to answer from sampled frames and audio in a single pass.
Open-source models following this approach include Qwen2.5-Omni~\citep{xu2025qwen25omnitechnicalreport}, Qwen3-Omni~\citep{xu2025qwen3}, MiniCPM-o~\citep{cui2026minicpm}, the video-SALMONN series~\citep{tang2025video}, LongCat-Flash-Omni~\citep{team2025longcat}, and Nemotron-Nano-Omni~\citep{deshmukh2026nemotron}.
Closed-source models such as Gemini, Qwen3.8-Omni-Flash~\citep{team2026qwen3}, and Qwen3.5-Omni-Plus~\citep{team2026qwen35} lead audio-visual benchmarks.
Progress in this direction relies heavily on scaling data and models, while acquiring new capabilities requires costly multimodal training.
Such training strengthens perception, while a single pass over fixed frame samples cannot revisit moments relevant to a question.
\method takes a complementary route that builds on the reasoning and tool use of a strong VLM, which accesses audio information through timestamped text and tools.

\noindent\textbf{Video Understanding Agent.}
Agents offer a promising approach to understanding long and complex videos~\citep{ren2026videorag,zhang2026deep,fan2024videoagent,xing2026native,tao2025active}.
Early methods reason over clip captions or decompose questions into steps handled by specialized tools, while recent agents actively decide where to look.
Audio-visual agents follow a similar progression: OmniAgent~\citep{tao2025active} equips o3 with tools backed by separate models; OmniReasoner~\citep{chen2026omnireasoner} and OmniAgent-7B~\citep{xing2026native} train omni models to use tools; and the Qwen3.8-Omni report evaluates agents in the Qwen Code harness.
TFO~\citep{deria2026training} provides a frozen VLM with frames and a speech transcript but answers in a single pass.
In contrast, \method equips a single VLM with general-purpose tools over video frames and a timestamped transcript, allowing it to decide where to look, how closely to inspect, and when to stop.
Concurrent agents take other routes: sandboxed coding agents~\citep{chen2026sandboxed} process videos using scripts and transcripts, and Omni-Decision~\citep{ma2026omni} combines a GPT-5.2 planner with Gemini-3.1-Pro perception, yet underperforms Gemini-3.1-Pro in a single pass.
\method exceeds Omni-Decision by 11.8 points on WorldSense, where it also leads every closed-source omni model (\cref{sec:appx_tables}).

%% file: sec/3_method.tex
\section{Method}
\label{sec:method}

\begin{figure*}[t]
  \centering
  \includegraphics[width=\linewidth]{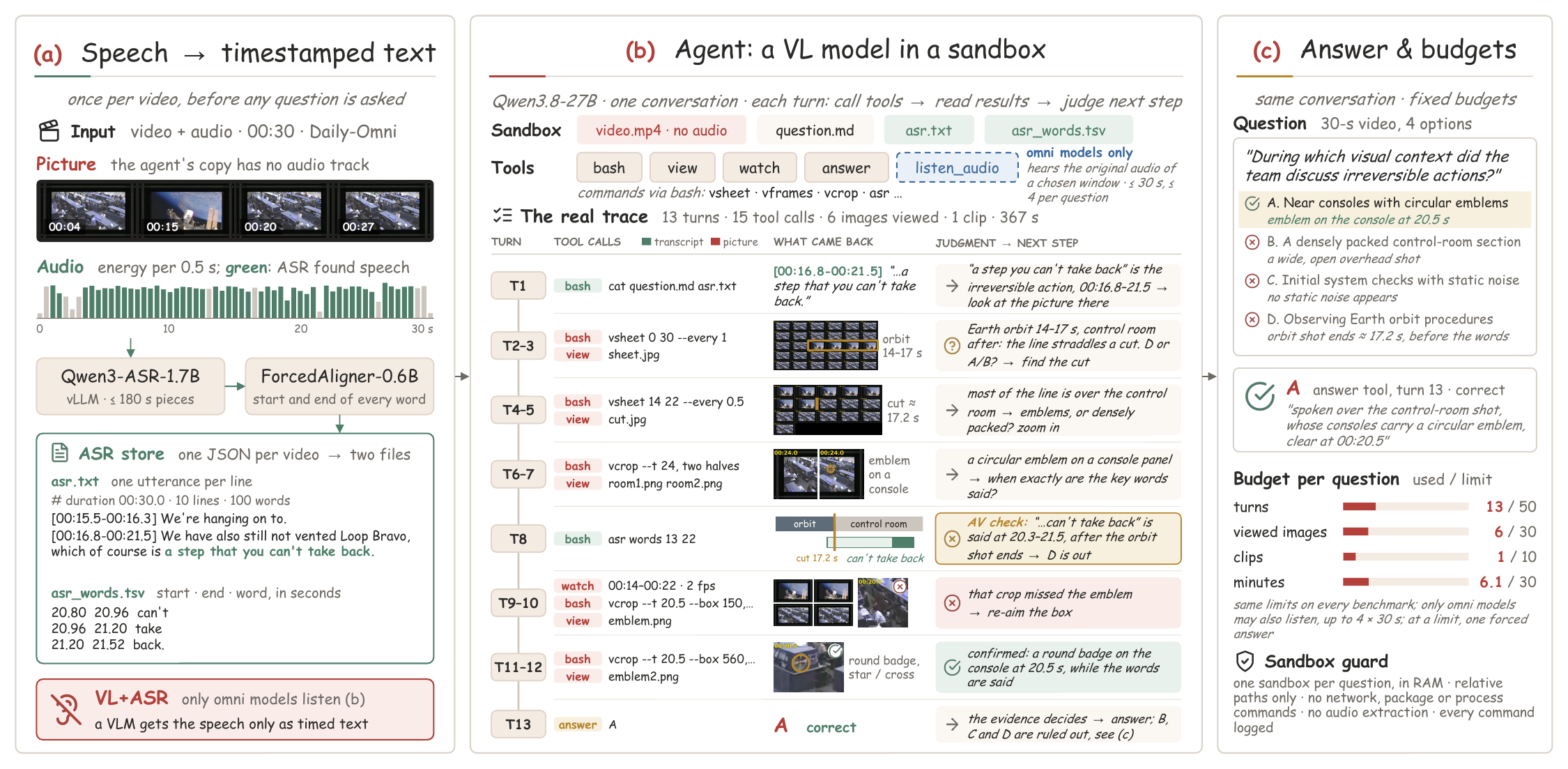}
  \caption{\textbf{Overview of \method.}
 (a)~Once per video, an ASR model transcribes the speech and a forced aligner assigns timestamps to each word, preserving non-speech intervals as gaps.
(b)~A VLM gathers evidence by calling tools over the muted video and transcript in a sandbox; only omni reasoning models also have access to \cmd{listen\_audio}.
(c)~The model answers once it has sufficient evidence, subject to fixed budgets.
The illustrated trajectory is a real example from Daily-Omni, answered correctly in 13 turns.}
  \label{fig:method}
\end{figure*}

\method answers audio-visual questions using a vision-language model (VLM) without direct access to audio. Instead, the audio is transcribed once per video with timestamps, and the VLM uses tools in a sandbox to gather evidence until it can answer (\cref{fig:method}).

%-------------------------------------------------------------------------
\subsection{Problem formulation}
\label{sec:method:problem}

Each example consists of a video of duration $D$, with frames $V$ and an audio track $A$, a question $q$, and a set of answer options $\mathcal{O}$.
An omni model processes both streams in a single turn:
\begin{equation}
  \hat{y} = \argmax_{o\in\mathcal{O}}\; p(o \mid V, A, q).
  \label{eq:omni}
\end{equation}
\method instead transcribes the audio once per video, independently of the question, to obtain $T=\mathrm{ASR}(A)$.
It then uses an agent built on a VLM $\pi$ to read $T$, inspect frames from $V$, and decide what evidence to gather next:
\begin{equation}
  \hat{y} = \mathrm{Agent}_{\pi}(V, T, q, \mathcal{O}).
  \label{eq:ours}
\end{equation}
The VLM receives no direct audio input and selects the evidence for each question autonomously.

%-------------------------------------------------------------------------
\subsection{Audio as timed text}
\label{sec:method:asr}

An ASR model transcribes the speech, and a forced aligner gives every word $w_j$ its start time $s_j$ and end time $e_j$ (\cref{fig:method}a):
\begin{equation}
  W = \{(w_j, s_j, e_j)\}_{j=1}^{M}.
  \label{eq:words}
\end{equation}
Stretches of at least $\gamma$ seconds without a word are kept as gaps,
\begin{equation}
  G = \{(e_j, s_{j+1}) : s_{j+1} - e_j \ge \gamma\},
  \label{eq:gaps}
\end{equation}
with $e_0=0$ and $s_{M+1}=D$.
The sandbox holds the transcript as two files: \cmd{asr.txt} lists the sentences and the gaps in time order, one line \cmd{[$s$-$e$] sentence} or \cmd{[no words $s$-$e$]} each, and \cmd{asr\_words.tsv} lists the words of $W$ with their times.
The transcript carries what was said and when, lyrics included, and its gaps mark the stretches where no one speaks, which often hold music or other sounds.

%-------------------------------------------------------------------------
\subsection{Agentic harness}
\label{sec:method:harness}

The harness consists of a scope definition, which fixes what the model can access and do, and an operating guide, which suggests how to gather evidence; both are part of the system prompt and are the same on every benchmark.

\paragraph{Scope definition.}
The files are the muted video, the question with its options, \cmd{asr.txt} and \cmd{asr\_words.tsv}.
The tools are \cmd{bash}, \cmd{view} (show images), \cmd{watch} (show a short clip) and \cmd{answer}.
Inside \cmd{bash}, \cmd{vsheet} renders a contact sheet of time-stamped frames over $[t_0,t_1]$, \cmd{vframes} extracts frames at a chosen rate, optionally at native resolution, \cmd{vcrop} crops a native-resolution frame, \cmd{asr} searches the transcript or lists word times and gaps, and standard tools such as \cmd{grep} and \cmd{python3} are available.
The rules keep the model inside its sandbox (relative paths, no network, no package or process commands, no audio extraction), and the budget $\Budget=(N_{\text{turn}},N_{\text{img}},N_{\text{clip}},T_{\max})$ limits the turns, viewed images, clips and time per question.

\paragraph{Operating guide.}
The guide suggests a sequence that usually works: read the question and the transcript, get an overview from one contact sheet, localize a window from transcript hits, sheet cells and gaps, zoom in, check the detail asked about, and answer once the evidence decides the question.
The model decides which steps a question needs.

\paragraph{Interaction.}
At turn $t$, the VLM reads its history $h_t$, which starts from the system prompt and the question, writes its reasoning and tool calls $a_t$, and receives their results $x_t$ (text, images or a clip):
\begin{equation}
  a_t \sim \pi(\cdot \mid h_t), \qquad h_{t+1} = h_t \oplus (a_t, x_t).
  \label{eq:loop}
\end{equation}
The trajectory ends when the model calls $\cmd{answer}(o)$, giving $\hat{y}=o$; when the budget is used up, the model answers with its most likely option.
Any VLM can serve as $\pi$ without other changes (\cref{tab:backbones}).

%-------------------------------------------------------------------------
\subsection{The transcript as a temporal index}
\label{sec:method:index}

The time stamps of the transcript tell the model where to look.
Looking through a video every $\Delta$ seconds costs $D/\Delta$ frames, 2100 for a 35-minute video at one frame per second.
Instead, the model searches the transcript for the words of the question, and a hit spoken over $[s,e]$ gives the window
\begin{equation}
  \Omega = [\,s-\delta,\; e+\delta\,]
  \label{eq:window}
\end{equation}
with a margin $\delta$ of its choosing; for a sound that is not speech, the gaps and scene changes give the candidates.
It then zooms in from coarse to fine, on windows $\Omega_1 \supset \Omega_2 \supset \cdots$ at intervals $\Delta_1 > \Delta_2 > \cdots$, and views
\begin{equation}
  n = \sum_k |\Omega_k| / \Delta_k \;\ll\; D / \Delta_K
  \label{eq:cost}
\end{equation}
frames in total.
The word times also tell whether a phrase and a visual event happen together: a phrase spoken over $[s,e]$ and an event seen over $[\alpha,\beta]$ coincide when
\begin{equation}
  [\,s,\, e\,] \cap [\,\alpha,\, \beta\,] \ne \varnothing,
  \label{eq:align}
\end{equation}
which is how the trajectory in \cref{fig:method}b rules out option D at turn 8.
The model takes these steps on its own; the harness prescribes no window or interval.

%-------------------------------------------------------------------------
\subsection{Listening on demand (omni models only)}
\label{sec:method:listen}

When the reasoning model is an omni model, the harness adds one tool, $\cmd{listen\_audio}(t_0,t_1)$, which returns the audio $A[t_0,t_1]$ as input, with $t_1-t_0\le L_{\text{a}}$ and at most $N_{\text{listen}}$ calls per question.
A VLM, such as the Qwen3.8-27B model of our main results, does not get this tool.

%% file: sec/4_experiments.tex
\section{Experiments}
\label{sec:exp}

\begin{table*}[t]
  \caption{Comparison with omni models and agents on four omni benchmarks (accuracy, \%).
  A+V: audio as input; V+ASR: speech as a timed transcript, no direct audio.
  \best{Bold}: best; \second{underline}: second best; \na: not reported; Avg.: mean over the four benchmarks.
  In parentheses: difference from the best open-source omni model (\textcolor{ForestGreen}{green}) or the best closed-source omni model or Qwen-code agent (\textcolor{red!80!black}{red}).}
  \label{tab:main}
  \centering
  \footnotesize
  \setlength{\tabcolsep}{2.3pt}
  \renewcommand{\arraystretch}{1.1}
  \begin{tabular}{llccccc}
    \toprule
    Model & Input & Daily-Omni & WorldSense & OmniVideoBench & LVOmniBench & Avg. \\
    \midrule
    \group{Closed-Source Omni Models} \\
    Gemini 3.8 Flash          & A+V & 84.0 & 69.6 & 65.2 & 70.7 & 72.4 \\
    Qwen3.8-Omni-Flash ~\citep{team2026qwen3}        & A+V & \best{85.1} & 68.5 & 63.4 & 63.3 & 70.1 \\
    Qwen3.5-Omni-Plus ~\citep{team2026qwen35}         & A+V & \best{85.1} & 63.9 & 53.8 & 53.2 & 64.0 \\
    \midrule
    \group{Open-Source Omni Models} \\
    Realtime-Venus-Omni ~\citep{group2026realtime}      & A+V & 81.3 & 54.0 & 39.2 & 38.9 & 53.4 \\
    MiniCPM-o 4.5 ~\citep{cui2026minicpm}             & A+V & 80.2 & 55.7 & 41.1 & 34.8 & 53.0 \\
    Qwen3-Omni-30B-A3B ~\citep{xu2025qwen3}        & A+V & 71.9 & 54.0 & 38.4 & 35.8 & 50.0 \\
    OmniVideo-R1 ~\citep{chen2026omnivideo}                               & A+V & 82.8 & 65.8 & 44.8 & \na & \na \\
    video-SALMONN-R3 ~\citep{li2026video}          & A+V & 78.7 & \na & 43.9 & 42.9 & \na \\
    LongCat-Flash-Omni ~\citep{team2025longcat}        & A+V & 82.4 & 60.9 & \na & \na & \na \\
    video-SALMONN 2+ (72B) ~\citep{tang2025video}    & A+V & 79.4 & 56.5 & \na & \na & \na \\
    Nemotron 3 Nano Omni~ \citep{deshmukh2026nemotron}      & A+V & 74.5 & 55.2 & \na & \na & \na \\
    \midrule
    \group{Omni Agents} \\
    OmniReasoner ~\citep{chen2026omnireasoner}              & A+V & 64.2 & 46.7 & 34.8 & 35.4 & 45.3 \\
    OmniAgent (o3 + Gemini + Qwen) ~\citep{tao2025active} & A+V & 82.7 & 61.2 & 59.1 & \na & \na \\
    OmniAgent (7B) ~\citep{xing2026native}            & A+V & 64.8 & 47.2 & 37.1 & \na & \na \\
    Qwen3.8-Omni-Flash(Qwen-code)       & A+V & \na & \na & 67.8 & 73.6 & \na \\
    Gemini 3.8 Flash(Qwen-code)         & A+V & \na & \na & \second{70.1} & 70.7 & \na \\
    \midrule
    \group{VL + ASR Agents} \\
    Qwen3-VL-30B-A3B + TFO ~\citep{deria2026training}    & V+ASR & 68.9 & 51.9 & \na & \na & \na \\
    \ours \method (Qwen3.8-27B, medium) & V+ASR & 80.3\dopen{$-$2.5} & 62.8\dopen{$-$3.0} & 58.3\dopen{+13.5} & 60.2\dopen{+17.3} & 65.4\dopen{+12.0} \\
    \ours \method (Qwen3.8-27B) & V+ASR & 80.4\dopen{$-$2.4} & 64.5\dopen{$-$1.3} & 60.4\dopen{+15.6} & 61.7\dopen{+18.8} & 66.8\dopen{+13.4} \\
    \ours \method (Opus 5.5)    & V+ASR & \second{84.4}\dclosed{$-$0.7} & \second{70.1}\dclosed{+0.5} & 69.9\dclosed{$-$0.2} & \second{75.5}\dclosed{+1.9} & \second{75.0}\dclosed{+2.6} \\
    \ours \method (Opus 5.5)    & V+ASR+A2M & \second{84.4}\dclosed{$-$0.7} & \best{71.1}\dclosed{+1.5} & \best{70.6}\dclosed{+0.5} & \best{76.2}\dclosed{+2.6} & \best{75.6}\dclosed{+3.2} \\
    \bottomrule
  \end{tabular}
\end{table*}

%-------------------------------------------------------------------------
\subsection{Setup}
\label{sec:exp:setup}

\paragraph{Benchmarks.}
We evaluate on four omni video question-answering benchmarks, spanning videos from 30 seconds to 90 minutes: Daily-Omni~\citep{zhou2025daily} (1,197 questions on 684 clips of 30 or 60\,s), WorldSense~\citep{hong2026worldsense} (3,172 questions on 1,662 videos, averaging 2.4\,min), OmniVideoBench~\citep{li2026omnivideobench} (1,000 questions on 628 videos, averaging 6.4\,min), and LVOmniBench~\citep{tao2026lvomnibench} (1,014 questions on 275 videos of 10--90\,min, averaging 34.6\,min).
We also evaluate on Video-MME-v2~\citep{fu2026video} (3,200 questions on 800 videos, averaging 10.4\,min, with questions grouped in sets of four), which reports average accuracy and a grouped nonlinear score across three difficulty levels.
All questions are multiple choice.

\paragraph{Compared methods.}
\cref{tab:main} compares \method with closed-source and open-source omni models, as well as omni and VL + ASR agents.
The agent baselines include OmniAgent, OmniReasoner, Qwen Code agents, and TFO.
We use scores reported in the corresponding papers or technical reports; all closed-source results are from the Qwen3.8-Omni technical report.

\paragraph{Implementation details.}
\label{sec:exp:impl}
The reasoning model is Qwen3.8-27B, served with vLLM on H200/A100 GPUs; \method (Opus 5.5) uses Claude Opus 5.5 instead and changes nothing else.
All reasoning models run with high reasoning effort; the ablations in \cref{sec:ablation} and the results marked medium use Qwen3.8-27B with medium effort.
The transcript is produced by Qwen3-ASR-1.7B and Qwen3-ForcedAligner-0.6B, with gaps of $\gamma=3$\,s.
On every benchmark, the budget $\Budget$ is 50 turns, 30 images, 10 clips and 30 minutes per question.
Omni reasoning models may also call \cmd{listen\_audio}, up to 4 times on windows of at most 30\,s.
More details are in \cref{sec:appx_impl}.

%-------------------------------------------------------------------------
\subsection{Main results}
\label{sec:exp:main}

\cref{tab:main} compares \method with frontier omni models and agents on the four benchmarks; \method sets a new state of the art both with a closed-source and with an open reasoning model.

\begin{figure*}[t]
  \centering
  \includegraphics[width=\textwidth]{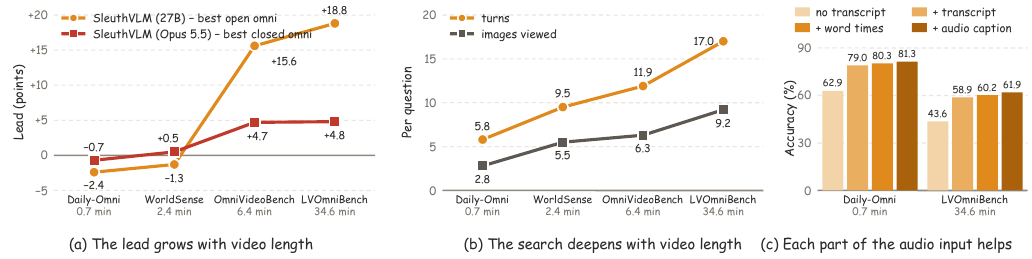}
  \caption{(a)~Lead of \method over the best omni model, with benchmarks ordered by average video length.
  (b)~Turns and images viewed per question by \method (Qwen3.8-27B, medium).
  (c)~Accuracy of \method (Qwen3.8-27B, medium) as each part of the audio input is added.}
  \label{fig:trends}
\end{figure*}

\paragraph{Comparison with omni models.}
Without hearing the audio, \method (Opus 5.5) achieves the best average accuracy, 75.0\%, above all closed-source omni models.
Without any training, \method (Qwen3.8-27B) is comparable to open-source omni models trained on large amounts of data: OmniVideo-R1 post-trains Qwen3-Omni-30B-A3B with RL on 101K samples using 128 GPUs, yet \method stays within 2.4 points on Daily-Omni and WorldSense and leads by 15.6 points on OmniVideoBench.
More training does not recover what a single pass over fixed frames misses in longer videos (\cref{fig:trends}a).

\paragraph{Comparison with agents.}
\method (Opus 5.5) outperforms OmniAgent, which coordinates o3 with three audio and video models, by up to 10.8 points, and \method (Qwen3.8-27B) surpasses the trained 7B agents by 15.6 to 26.3 points without any training.
The Qwen team has also moved towards agentic omni understanding with its Qwen Code harness; without hearing the audio directly, \method (Opus 5.5) is 2.1 points above its Qwen3.8-Omni-Flash agent and on par with its Gemini 3.8 Flash agent on OmniVideoBench, and ahead of both on LVOmniBench.
TFO, which also gives a VLM a transcript but answers in a single pass, trails \method (Qwen3.8-27B) by about 12 points: the transcript makes the audio accessible, and the agent turns it into accuracy.

\begin{table}[t]
  \caption{\method (Opus 5.5) achieves the best average over the four omni benchmarks and Video-MME-v2 (accuracy, \%).
  DO: Daily-Omni; WS: WorldSense; OVB: OmniVideoBench; LV: LVOmniBench; VM2: Video-MME-v2; Avg.: mean over the five benchmarks.
  QC: the model run as an agent in the Qwen Code harness.
  \best{Bold}: best; \second{underline}: second best; \na: not reported.}
  \label{tab:five}
  \centering
  \footnotesize
  \setlength{\tabcolsep}{2.6pt}
  \renewcommand{\arraystretch}{1.1}
  \begin{tabular}{lcccccc}
    \toprule
    Model & DO & WS & OVB & LV & VM2 & Avg. \\
    \midrule
    \ours \method (Opus 5.5) & \second{84.4} & \best{70.1} & \second{69.9} & \best{75.5} & \best{77.8} & \best{75.5} \\
    Gemini 3.8 Flash & 84.0 & \second{69.6} & 65.2 & 70.7 & 71.0 & \second{72.1} \\
    Qwen3.8-Omni-Flash & \best{85.1} & 68.5 & 63.4 & 63.3 & 65.0 & 69.1 \\
    \ours \method (Qwen3.8-27B) & 80.4 & 64.5 & 60.4 & 61.7 & 62.0 & 65.8 \\
    Qwen3.5-Omni-Plus & \best{85.1} & 63.9 & 53.8 & 53.2 & 47.9 & 60.8 \\
    \midrule
    Gemini 3.8 Flash (QC) & \na & \na & \best{70.1} & 70.7 & \second{72.7} & \na \\
    Qwen3.8-Omni-Flash (QC) & \na & \na & 67.8 & \second{73.6} & 71.3 & \na \\
    \bottomrule
  \end{tabular}
\end{table}

\begin{table*}[t]
  \caption{LVOmniBench accuracy (\%) by difficulty and ability.
V+ASR+Audio denotes omni models with access to \cmd{listen\_audio}.
Ability groupings and per-type results appear in \cref{sec:appx_tables,sec:appx_backbones}, respectively.
\best{Bold} and \second{underline} mark the best and second-best results; question counts are in parentheses.}
  \label{tab:backbones}
  \centering
  \small
  \setlength{\tabcolsep}{4pt}
  \renewcommand{\arraystretch}{1.1}
  \begin{tabular}{llcccccccc}
    \toprule
    & & \multicolumn{3}{c}{Difficulty} & \multicolumn{4}{c}{Ability} & \\
    \cmidrule(lr){3-5} \cmidrule(lr){6-9}
    Model & Input & Low & Medium & High & Underst. & Percept. & Infer. & Logical & Overall \\
          &       & \scriptsize(314) & \scriptsize(441) & \scriptsize(259) & \scriptsize(230) & \scriptsize(386) & \scriptsize(289) & \scriptsize(109) & \scriptsize(1014) \\
    \midrule
    \method (Opus 5.5)           & V+ASR       & \second{82.5} & \best{76.0} & \second{66.4} & \best{80.4} & \second{74.4} & \second{70.6} & \best{82.6} & \best{75.5} \\
    \method (Qwen3.8-Omni-Flash) & V+ASR+Audio & 80.6 & 71.4 & 57.1 & 73.9 & 68.7 & \best{71.6} & 67.9 & 70.6 \\
    \method (Gemini 3.8 Flash)   & V+ASR+Audio & \best{82.8} & \second{72.6} & \best{67.2} & \second{77.4} & \best{74.6} & \second{70.6} & \second{77.1} & \second{74.4} \\
    \midrule
    Qwen3.8-Omni-Flash           & A+V & \na & \na & \na & \na & \na & \na & \na & 63.3 \\
    Gemini 3.8 Flash             & A+V & \na & \na & \na & \na & \na & \na & \na & 70.7 \\
    Qwen3.5-Omni-Plus            & A+V & \na & \na & \na & \na & \na & \na & \na & 53.2 \\
    \bottomrule
  \end{tabular}
\end{table*}

\paragraph{Video-MME-v2.}
\method (Opus 5.5) also sets a new state of the art on Video-MME-v2: with 77.8\% average accuracy and a non-linear score of 63.5, it is the best model on every metric of the leaderboard, and over the five benchmarks it achieves the best average, 75.5\% (\cref{tab:five}).
It is 6.8 points above Gemini 3.8 Flash and 11.7 above Gemini-3-Pro in accuracy, 14.1 points above Gemini-3-Pro in non-linear score, and ahead of the Qwen Code agents of Gemini 3.8 Flash and Qwen3.8-Omni-Flash as well.
With Qwen3.8-27B, \method reaches 62.0\% accuracy and a non-linear score of 44.4, ahead of Gemini-3-Flash, GPT-5 and every open-source model on the leaderboard.
The full leaderboard is given in \cref{sec:appx_vmme2}.

\paragraph{Insight 1: Searching outperforms passive observation.}
Fixed sampling limits performance: when used as agents, the same omni models gain up to 10.3 points (\cref{tab:main,tab:backbones}).
\method also adapts its search to each video: with medium effort, it uses an average of 5.8 turns per question on Daily-Omni and 17.0 on LVOmniBench (\cref{fig:trends}b).
A strong base model needs only a simple but effective harness. Earlier agents delegate perception to task-specific tools backed by other models; OmniAgent, for example, uses seven such tools.
In contrast, the \method harness specifies the resources and budget available to the VLM while leaving perception decisions to the model itself (\cref{sec:method:harness}).
With this harness, a single VLM without direct audio access outperforms both OmniAgent and the Qwen Code agent based on Qwen3.8-Omni-Flash.

\begin{table}[t]
  \caption{LVOmniBench questions by question type and audio type, with the accuracy (\%) of \method (Qwen3.8-27B, medium) in parentheses.
  Non-speech: share of Music and Sound questions in the type.}
  \label{tab:lv_types}
  \centering
  \footnotesize
  \setlength{\tabcolsep}{2.5pt}
  \renewcommand{\arraystretch}{1.1}
  \begin{tabular}{lcccc}
    \toprule
    Type & Speech & Music & Sound & Non-speech \\
    \midrule
    Attribute Perception        & 145 (63) & 3 (67)  & 6 (67)  & 5.8\% \\
    Counting                    & 120 (60) & 11 (55) & 12 (42) & 16.1\% \\
    Spatial Inference           & 122 (66) & 5 (60)  & 2 (50)  & 5.4\% \\
    Event Understanding         & 115 (68) & 7 (43)  & 5 (80)  & 9.4\% \\
    Logical Reasoning           & 94 (61)  & 10 (70) & 5 (60)  & 13.8\% \\
    Human-Centric Underst. & 87 (60)  & 2 (50)  & 14 (43) & 15.5\% \\
    Temporal Inference          & 67 (60)  & 4 (25)  & 8 (50)  & 15.2\% \\
    Music Perception            & \na        & 89 (51) & \na       & 100\% \\
    Sound Inference             & 13 (54)  & 6 (83)  & 62 (52) & 84.0\% \\
    \midrule
    Total                       & 763 (63) & 137 (53) & 114 (52) & 24.8\% \\
    \bottomrule
  \end{tabular}
\end{table}

% case-study figure of \cref{sec:cases}, defined here so that it floats to the pages before it
\begin{figure*}[!tbp]
  \centering
  \includegraphics[width=0.95\textwidth]{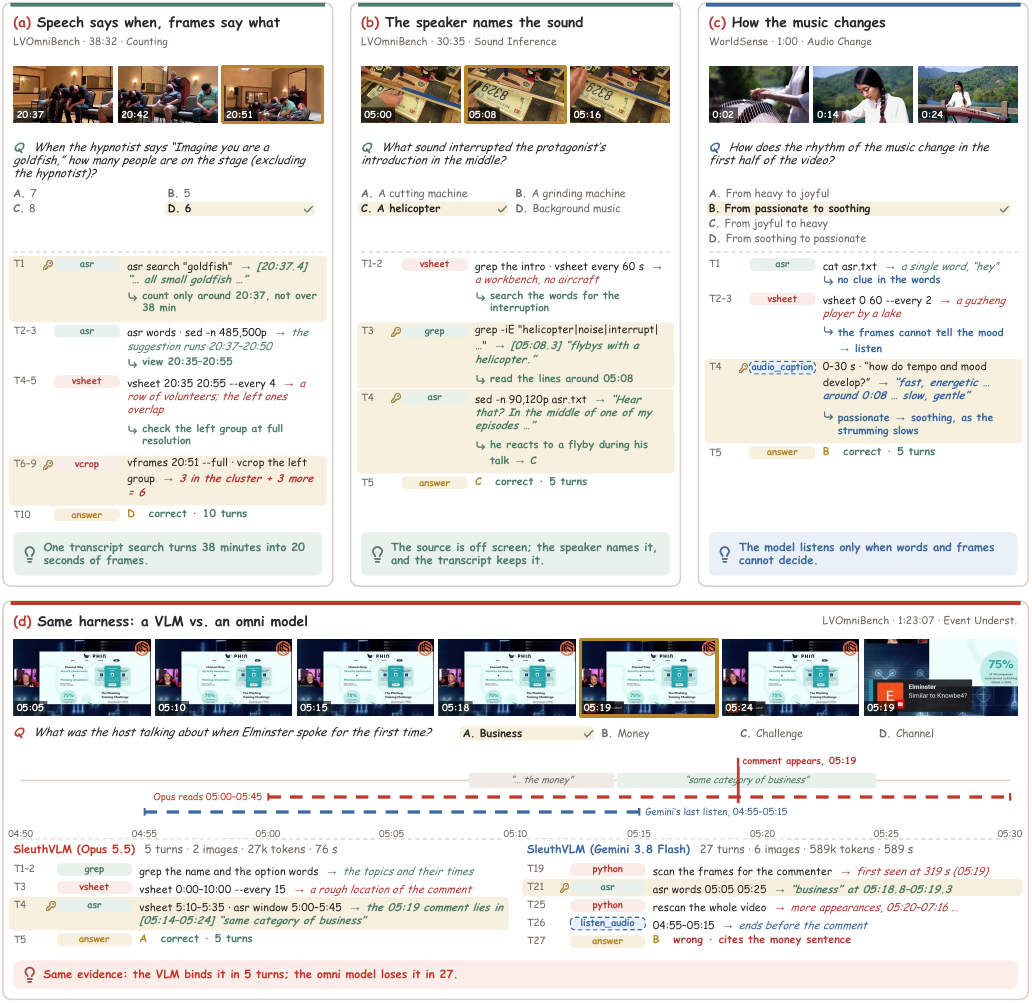}
  \caption{Four trajectories of \method: (a,~b)~answered from the transcript and the picture, (c)~with the \cmd{audio\_caption} tool, (d)~one question with Claude Opus 5.5, a VLM, and Gemini 3.8 Flash, an omni model, as the reasoning model.
  Highlighted rows hold the decisive evidence, and the line under a call is the model's decision; \textcolor[HTML]{4F7F6A}{green}: transcript, \textcolor[HTML]{B3403A}{red}: picture, \textcolor[HTML]{3D6A9E}{blue}: audio.
  Every step of (a--d) is detailed in \cref{sec:appx_cases}.}
  \label{fig:cases}
\end{figure*}

\paragraph{Insight 2: Agentic ability matters as much as direct audio access.}
Agentic inference improves omni understanding even when a model can listen directly: on LVOmniBench, our harness raises Gemini 3.8 Flash from 70.7\% to 74.4\% and Qwen3.8-Omni-Flash from 63.3\% to 70.6\% (\cref{tab:backbones}).
A VLM also performs strongly without direct audio access: given video and timestamped transcripts, \method with Opus 5.5 reaches 75.5\%, surpassing both omni models in the same harness.
Its largest lead over Gemini is on logical reasoning, at 82.6\% versus 77.1\%.
In one case, Opus connects a visual event to the corresponding transcript in five turns, while Gemini finds both but fails to connect them (\cref{sec:cases:agentic}).
Transcripts provide useful audio information because 75.2\% of LVOmniBench questions involve speech (\cref{tab:lv_types}); an external audio tool can further help a VLM understand non-speech audio (\cref{sec:ablation}).
These results highlight the value of agentic methods for offline understanding and suggest that agentic ability should be preserved during modality alignment as omni models scale.

\paragraph{Insight 3: Small agentic models are highly competitive.}
\method with Qwen3.8-27B runs on a single GPU, yet outperforms the closed-source Qwen3.5-Omni-Plus on average and performs within the range of closed-source omni models (\cref{fig:teaser}a).
The same harness also scales with the reasoning model: Opus 5.5 adds another 8.2 points.
A small model with strong agentic capabilities already matches closed-source omni models, highlighting the value of developing agentic ability alongside scaling.

\begin{table}[t]
  \caption{Ablations of \method (Qwen3.8-27B, medium) on a short-video (Daily-Omni, 30--60\,s) and a long-video benchmark (LVOmniBench, 10--90\,min), accuracy (\%) with the change from the full system A0 in parentheses.}
  \label{tab:ablation}
  \centering
  \small
  \setlength{\tabcolsep}{4pt}
  \renewcommand{\arraystretch}{1.1}
  \begin{tabular}{lcc}
    \toprule
    Setting & Daily-Omni & LVOmniBench \\
    \midrule
    \ours A0\; Full system      & 80.3 & 60.2 \\
    A1\; w/o ASR transcript     & 62.9 \textcolor{red!80!black}{\scriptsize($-$17.4)} & 43.6 \textcolor{red!80!black}{\scriptsize($-$16.6)} \\
    A2\; w/o word-level times   & 79.0 \textcolor{red!80!black}{\scriptsize($-$1.3)}  & 58.9 \textcolor{red!80!black}{\scriptsize($-$1.3)} \\
    \bottomrule
  \end{tabular}
\end{table}

%-------------------------------------------------------------------------
\subsection{Ablations}
\label{sec:ablation}

\paragraph{Timed transcript and word times.}
All ablations use \method (Qwen3.8-27B) at medium effort (A0: the medium row of \cref{tab:main}).
\cref{tab:ablation} removes the transcript (A1: only the muted video and the question remain) or only its word times (A2), and \cref{fig:trends}c adds the parts back one at a time.
Each part adds accuracy on both benchmarks.
Without the transcript, accuracy falls by 17.4 and 16.6 points, although on LVOmniBench the model takes 24.6 turns and views 15.1 images per question instead of 17.0 and 9.2; even its sound questions lose 10.5 points, as the transcript also tells the model where to look (\cref{sec:method:index}).
The word times add 1.3 points on both benchmarks at no extra turns, mostly on the music and sound questions of LVOmniBench (6.6 and 6.1 points), whose timing is read from the boundaries of the words around them.
\paragraph{External audio tools.}
According to \cref{tab:main} we find music and environmental sounds difficult to resolve,
so we add two tool useful audio tool: Audio to Math (A2M) extracts signal measurements and plots; Audio Caption provides textual descriptions.
On WorldSense and OmniVideoBench, Music accuracy rises by 3.2 and 2.2 points with A2M (Opus 5.5), and by 10.9 and 2.2 with Audio Caption (Qwen3.8-27B), respectively.
By recovering acoustic evidence omitted by ASR (\cref{sec:appx_audiocap,sec:appx_audiomath}), these tools enable strong VLMs to excel at omni tasks, underscoring the importance of preserving foundational capabilities when scaling omni-modal models.

%% file: sec/5_cases.tex
\FloatBarrier
\section{Case Studies}
\label{sec:cases}

\cref{fig:cases} shows four trajectories with the decisive steps highlighted; \cref{sec:appx_cases} walks through each of them turn by turn.

%-------------------------------------------------------------------------
\subsection{Answering from the timed transcript}
\label{sec:cases:asr}

In \cref{fig:cases}a, one search of the transcript lands at 20:37 of a 38-minute show, and the model counts six people after viewing only the 20 seconds around it.
In \cref{fig:cases}b, the helicopter never appears on screen, but the transcript holds the speaker naming it: ``flybys with a helicopter. Hear that?''.
The transcript gives the decisive clue, a moment in (a) and the sound itself in (b), and the frames confirm it.

%-------------------------------------------------------------------------
\subsection{Hearing the music through a tool}
\label{sec:cases:caption}

In \cref{fig:cases}c, neither the transcript, a single word, nor the frames tell how the music changes.
One call to \cmd{audio\_caption} on the first 30\,s returns a fast opening that turns slow and gentle near 0:08, and the model checks it against the strumming in the frames.

%-------------------------------------------------------------------------
\subsection{Agentic ability: a VLM and an omni model in one harness}
\label{sec:cases:agentic}

In \cref{fig:cases}d, Claude Opus 5.5, a VLM, and Gemini 3.8 Flash, an omni model that may also listen, answer the same question in the same harness: what was the host saying when a viewer first commented, at 05:19 of an 83-minute stream?
The key step is to bind the comment to the sentence spoken at that moment, ``same category of business''.
Opus 5.5 reads the frames and the transcript of the same window and binds them in five turns.
Gemini 3.8 Flash finds the comment at 05:19 and the word ``business'' at 05:18.8, then rescans the whole video, listens to a window that ends before the comment, and answers Money after 27 turns and 21 times the tokens.
This illustrates Insight~2: once the evidence is found, planning and evidence binding decide the question, and here the VLM binds it in one turn.

%% file: sec/6_conclusion.tex
\section{Conclusion}
\label{sec:conclusion}

We explored omni video understanding without direct audio access and introduced \method, an agent harness that enables a VLM to read a word-timestamped transcript and autonomously search the video.
\method establishes a new state of the art across four omni benchmarks and on Video-MME-v2.
With Opus 5.5, it achieves an average accuracy of 75.0\% across the four benchmarks, outperforming all closed-source omni model, and leads the Video-MME-v2 leaderboard with 77.8\%.
With Qwen3.8-27B, it is comparable to open-source omni models on short videos and better on long ones.
Our results highlight three insights: 1.\textit{Searching outperforms passive observation}, 2. \textit{Agentic ability matters as much as direct audio access}, and 3. \textit{Small agentic models are highly competitive}.
\textbf{\method provides a simple but effective starting point for doing omni video understanding.}

%% file: sec/appendix.tex
The appendix follows the order of the paper.
\cref{sec:appx_impl} gives the implementation details of \cref{sec:exp:setup}, from the transcript to the sandbox, together with an audit of every trajectory.
\cref{sec:appx_tables,sec:appx_backbones,sec:appx_audiocap} expand the results on the four omni benchmarks: the full leaderboards behind \cref{tab:main}, the reasoning models of \cref{tab:backbones} by question type and cost, and the audio caption tool of \cref{sec:ablation} on the music questions.
\cref{sec:appx_vmme2} lists the full Video-MME-v2 leaderboard behind \cref{tab:five}.
\cref{sec:appx_cases} walks through the trajectories of \cref{sec:cases} turn by turn.
\cref{sec:appx_audiomath} details the Audio to Math tools and walks through a tuning-interval question resolved by time-resolved frequency measurements (\cref{fig:appx_audiomath_case}).

%-------------------------------------------------------------------------
\section{Implementation details}
\label{sec:appx_impl}
This section complements \cref{sec:exp:setup} and follows the pipeline of \method: each video is transcribed once, and the reasoning model then works in a sandbox with a fixed set of tools.

\begin{table*}[t]
  \caption{Full comparison on Daily-Omni (1197 questions) by question type and video duration (accuracy, \%).
  Baselines are the ``With Visual and Audio'' rows of the Daily-Omni paper; both the Instruct and Thinking versions of Qwen3-Omni are listed.
  Overall scores only: $^{\dagger}$ from the video-SALMONN~2 paper, $^{\S}$ from the TFO paper, $^{\P}$ from the video-SALMONN-R3 paper.}
  \label{tab:appx_daily}
  \centering
  \footnotesize
  \setlength{\tabcolsep}{3pt}
  \renewcommand{\arraystretch}{1.05}
  \begin{tabular}{llccccccccc}
    \toprule
    Model & Input & AV Align & Comp. & Ctx.\ Und. & Evt.\ Seq. & Infer. & Reas. & 30\,s & 60\,s & Avg. \\
    \midrule
    \ours \method (Opus 5.5) & V+ASR & \second{81.51} & \second{86.26} & \second{84.97} & \best{82.35} & \best{86.36} & \best{88.00} & \best{83.15} & \second{85.82} & \best{84.38} \\
    \ours \method (Opus 5.5) & V+ASR+A2M & \best{81.93} & \best{87.02} & \best{87.05} & \best{82.35} & 83.77 & \second{86.86} & \second{83.00} & \best{86.00} & \best{84.38} \\
    \ours \method (Qwen3.8-27B, medium) & V+ASR+Audio Caption & 79.83 & 85.50 & 80.31 & 77.78 & 84.42 & 84.57 & 80.06 & 82.73 & \second{81.29} \\
    \ours \method (Qwen3.8-27B) & V+ASR & 76.89 & 81.68 & 78.76 & 79.08 & 83.77 & 85.14 & 77.28 & 84.00 & 80.37 \\
    \ours \method (Qwen3.8-27B, medium) & V+ASR & 77.31 & 80.92 & 76.68 & \second{80.39} & 84.42 & 84.00 & 78.05 & 82.91 & 80.28 \\
    video-SALMONN 2+ (72B) & Open, A+V & \na & \na & \na & \na & \na & \na & \na & \na & 79.4$^{\dagger}$ \\
    video-SALMONN-R3 (8B) & Open, A+V & \na & \na & \na & \na & \na & \na & \na & \na & 78.7$^{\P}$ \\
    Qwen3-Omni-Flash & Closed, A+V & \na & \na & \na & \na & \na & \na & \na & \na & 76.2$^{\dagger}$ \\
    Qwen3-Omni-30B-A3B-Thinking & Open, A+V & 65.97 & 80.92 & 65.80 & 71.57 & \second{85.06} & 80.57 & 76.04 & 70.73 & 73.60 \\
    Gemini 2.5 Flash & Closed, A+V & 73.82 & 66.41 & 72.04 & 68.03 & 78.67 & 81.87 & 69.86 & 77.09 & 73.06 \\
    Qwen3-Omni-30B-A3B-Instruct & Open, A+V & 66.81 & 80.92 & 64.77 & 66.34 & 81.17 & 81.14 & 71.87 & 71.82 & 71.85 \\
    video-SALMONN 2+ (7B) & Open, A+V & \na & \na & \na & \na & \na & \na & \na & \na & 71.8$^{\dagger}$ \\
    Qwen3-VL-30B-A3B + TFO & V+ASR & \na & \na & \na & \na & \na & \na & \na & \na & 68.9$^{\S}$ \\
    Gemini 2.0 Flash & Closed, A+V & 62.18 & 73.28 & 63.73 & 63.72 & 76.62 & 75.43 & 67.23 & 68.55 & 67.84 \\
    video-SALMONN 2+ (3B) & Open, A+V & \na & \na & \na & \na & \na & \na & \na & \na & 67.7$^{\dagger}$ \\
    video-SALMONN 2 F-16 (7B) & Open, A+V & \na & \na & \na & \na & \na & \na & \na & \na & 66.8$^{\dagger}$ \\
    video-SALMONN 2 (7B) & Open, A+V & \na & \na & \na & \na & \na & \na & \na & \na & 66.3$^{\dagger}$ \\
    Qwen2.5-Omni-7B & Open, A+V & 48.32 & 69.47 & 58.55 & 58.17 & 76.62 & 73.14 & 64.61 & 59.09 & 62.07 \\
    Gemini 2.5 Flash Lite & Closed, A+V & 57.56 & 68.70 & 56.48 & 52.61 & 79.22 & 69.71 & 63.52 & 60.00 & 61.90 \\
    Daily-Omni-Baseline-Qwen2.5 & Open, A+V & 51.68 & 68.70 & 60.10 & 53.92 & 78.57 & 71.43 & 63.99 & 59.27 & 61.82 \\
    Gemini 2.0 Flash Lite & Closed, A+V & 55.04 & 64.89 & 58.03 & 54.25 & 74.03 & 72.00 & 62.44 & 60.00 & 61.32 \\
    Qwen2.5-Omni-3B & Open, A+V & 50.84 & 69.47 & 53.89 & 53.92 & 75.97 & 70.29 & 62.60 & 57.45 & 60.23 \\
    \ours \method (Qwen3-Omni-30B-A3B) & V+ASR+Audio & 53.78 & 48.09 & 54.40 & 42.48 & 64.94 & 67.43 & 55.33 & 52.00 & 53.80 \\
    Ola (7B) & Open, A+V & 40.34 & 61.07 & 40.41 & 43.46 & 63.64 & 69.71 & 51.47 & 49.82 & 50.71 \\
    VideoLLaMA2 (7B) & Open, A+V & 35.71 & 35.88 & 35.75 & 31.70 & 40.91 & 34.29 & 38.02 & 31.82 & 35.17 \\
    Unified-IO-2 XL (3B) & Open, A+V & 30.25 & 30.53 & 25.39 & 29.08 & 33.12 & 21.71 & 28.13 & 28.55 & 28.32 \\
    Unified-IO-2 XXL (8B) & Open, A+V & 25.63 & 31.30 & 26.42 & 25.82 & 35.06 & 29.71 & 26.74 & 30.00 & 28.24 \\
    Unified-IO-2 L (1B) & Open, A+V & 27.31 & 22.90 & 26.42 & 27.78 & 29.87 & 29.14 & 27.67 & 27.09 & 27.40 \\
    \bottomrule
  \end{tabular}
\end{table*}

\paragraph{Transcription.}
Long recordings are cut at quiet points into pieces of at most 180\,s, transcribed and aligned piecewise, and merged back on the video's time axis.
Languages outside the aligner's coverage are transcribed in 30\,s chunks with chunk-level times.
Transcribing the 3249 videos of the four benchmarks, about 302 hours of audio, takes about 113 minutes on a single H200.

\paragraph{Tools and commands.}
\cmd{view} shows up to four images per call, and a contact sheet counts as one image;
\cmd{watch} shows a segment of at most 30\,s at 1 to 8\,fps and at most 32 frames;
\cmd{vsheet} tiles up to 64 time-stamped frames, \cmd{vframes} extracts up to 32 frames, and \cmd{vcrop} takes its box on a 0--1000 scale of a native-resolution frame.
\cmd{listen\_audio}, offered to omni reasoning models only, returns a mono 16\,kHz clip of the original audio.
Each command may run for 120\,s, and at most 8000 characters of its output are kept in the conversation.

\paragraph{Sandbox.}
Each question gets its own sandbox in memory, holding a muted copy of the video, \cmd{question.md} with the question and its options, and the two transcript files.
Commands run with the sandbox as working and home directory and a minimal search path.
Before a command runs, a guard checks it against fixed rules and refuses absolute paths to system directories, climbing out with \cmd{..}, network, package, process and system commands such as \cmd{curl}, \cmd{wget}, \cmd{ssh}, \cmd{git}, \cmd{pip} and \cmd{sudo}, the Python modules for network and process control, and extracting audio with \cmd{ffmpeg}.
A refused command is not run, and the model is told the reason.
Besides \cmd{bash}, \cmd{view} opens only image files inside the sandbox, \cmd{watch} plays only the sandbox video, and \cmd{answer} ends the question.
Every command and its output are written to the result file, and every refused command is kept with its reason.

\paragraph{Tools are shortcuts, not limits.}
The harness gives the model a shell rather than a fixed set of operations.
When \cmd{watch} and \cmd{vcrop} are withheld, \method (Qwen3.8-27B, medium) rebuilds both on Daily-Omni: it extracts native-resolution frames with \cmd{vframes} and crops and enlarges regions with PIL, and it samples a short window densely and tiles the frames into its own time-labeled sheets to look at a moment again.
In a free-form harness, the model decides not only where to look but also how, and a missing tool becomes a few lines of code.

\paragraph{Reasoning models and serving.}
Qwen3.8-27B runs in thinking mode with high reasoning effort (medium for the ablations and the results marked medium), sampled at temperature 1.0, top-$p$ 0.95 and top-$k$ 20, with at most 8192 generated tokens per turn (reasoning included) and up to four tool calls per turn.
It is served with vLLM on one H200 GPU per shard, with sixteen questions in flight per GPU; a finished question is replaced by the next one at once.
Earlier reasoning stays in the history, so prompts only grow and the prefix cache is reused; when a prompt exceeds 110k tokens, the oldest images are dropped from it.
Claude Opus 5.5, Qwen3.8-Omni-Flash and Gemini 3.8 Flash are called through APIs with the same harness, prompts and budget; time spent waiting on rate limits does not count towards $T_{\max}$.

\paragraph{Trajectory audit.}
We audit every trajectory and log of \method (Qwen3.8-27B, medium) on the four benchmarks, 6,383 questions and 71,132 tool calls, including every command that builds a path at run time.
Every answer is drawn from the video, the question and the transcript: no command opens an annotation or result file, no output contains an answer, and the guard refused every network command.

%-------------------------------------------------------------------------
\section{Full results on the four omni benchmarks}
\label{sec:appx_tables}
\cref{tab:main} reports one score per benchmark; \cref{tab:appx_daily,tab:appx_worldsense,tab:appx_ovb,tab:appx_lvomni} give the full leaderboards with every category each benchmark reports.
\method (Opus 5.5) leads every category in all four tables.
With Qwen3.8-27B, \method leads the strongest open-source omni model by 13.4 points on average (\cref{tab:main}) and by 18.8 points on LVOmniBench.
\method also leads the agents: with Opus 5.5 it exceeds Omni-Decision, a GPT-5.2 planner over Gemini-3.1-Pro perception, by 11.8 points on WorldSense, and with Qwen3.8-27B it exceeds TFO, which gives a VLM the same kind of transcript in a single pass, by 11.5 points on Daily-Omni and 12.6 on WorldSense.

The three \method (Qwen3.8-27B) rows are the main setting and the medium-effort setting of the ablations in \cref{sec:ablation}, without and with the audio caption tool (\cref{sec:appx_audiocap}).
Rows are sorted by the overall score and our rows are shaded; \best{bold}: best, \second{underline}: second best, \na: not reported.
Scores of other methods are as reported by their authors.

The \method (Qwen3-Omni-30B-A3B) rows use Qwen3-Omni-30B-A3B-Instruct as the reasoning model in the same harness, where it can also call \cmd{listen\_audio}.
The model hears the audio but lacks agentic ability, and agentic reasoning collapses: it scores 53.8, 38.0, 31.6 and 27.9\% on the four benchmarks, 12.2 points below its own single-pass average (\cref{tab:main}).
With stronger agentic ability, Qwen3.8-Omni-Flash and Gemini 3.8 Flash gain 7.3 and 3.7 points from the same harness on LVOmniBench (\cref{tab:backbones}).
Agentic ability is therefore as important as omni scale.

\begin{table*}[t]
  \caption{Full comparison on WorldSense (3172 questions) by domain (accuracy, \%): Tech \& Science, Culture \& Politics, Daily Life, Film \& TV, Performance, Games, Sports, Music.
  Baselines are the audio-visual models of the WorldSense paper; Omni-Decision, a GPT-5.2 planner over Gemini-3.1-Pro perception, is from its paper.
  Overall scores only: $^{\dagger}$ from the video-SALMONN~2 paper, $^{\S}$ from the TFO paper.}
  \label{tab:appx_worldsense}
  \centering
  \footnotesize
  \setlength{\tabcolsep}{3.5pt}
  \renewcommand{\arraystretch}{1.05}
  \begin{tabular}{llccccccccc}
    \toprule
    Model & Input & Tech. & Culture & Daily & Film\,\&\,TV & Perf. & Games & Sports & Music & Avg. \\
    \midrule
    \ours \method (Opus 5.5) & V+ASR+A2M & \best{73.1} & \best{72.5} & \best{69.9} & \best{76.3} & \best{74.5} & \second{67.0} & \best{68.4} & \best{68.0} & \best{71.1} \\
    \ours \method (Opus 5.5) & V+ASR & \second{71.2} & \second{72.2} & \best{69.9} & \second{74.9} & \second{72.7} & \best{71.2} & \second{66.0} & \second{64.8} & \second{70.1} \\
    Gemini 2.5 Pro & Closed, A+V & 64.9 & 66.0 & 65.8 & 68.1 & 69.7 & 65.7 & 63.5 & 61.3 & 65.1 \\
    \ours \method (Qwen3.8-27B) & V+ASR & 69.6 & 69.9 & \second{66.6} & 67.8 & 66.7 & 65.7 & 56.5 & 53.9 & 64.5 \\
    \ours \method (Qwen3.8-27B, medium) & V+ASR+Audio Caption & 70.2 & 68.6 & 61.9 & 66.0 & 65.9 & 63.9 & 57.9 & 63.1 & 64.4 \\
    \ours \method (Qwen3.8-27B, medium) & V+ASR & 69.6 & 67.6 & 60.8 & 67.0 & 65.9 & 62.7 & 59.3 & 52.2 & 62.8 \\
    Omni-Decision & Agent, A+V & 63.7 & 60.5 & 53.3 & 66.8 & 52.4 & 56.7 & 50.0 & 64.0 & 58.3 \\
    video-SALMONN 2+ (72B) & Open, A+V & 59.0 & 63.1 & 54.0 & 59.9 & 58.1 & 54.1 & 51.9 & 54.4 & 56.5 \\
    Qwen3-Omni-Flash & Closed, A+V & \na & \na & \na & \na & \na & \na & \na & \na & 54.1$^{\dagger}$ \\
    Qwen3-Omni & Open, A+V & 58.7 & 60.5 & 54.5 & 53.8 & 55.4 & 46.8 & 48.8 & 52.2 & 54.0 \\
    Gemini 2.5 Flash & Closed, A+V & 51.8 & 50.2 & 54.1 & 51.2 & 59.6 & 50.6 & 51.6 & 51.5 & 52.3 \\
    Qwen3-VL-30B-A3B + TFO & V+ASR & \na & \na & \na & \na & \na & \na & \na & \na & 51.9$^{\S}$ \\
    video-SALMONN 2+ (7B) & Open, A+V & 57.1 & 54.4 & 48.9 & 50.9 & 49.1 & 51.1 & 44.9 & 51.0 & 50.9 \\
    video-SALMONN 2 F-16 (7B) & Open, A+V & \na & \na & \na & \na & \na & \na & \na & \na & 49.6$^{\dagger}$ \\
    video-SALMONN 2 (7B) & Open, A+V & \na & \na & \na & \na & \na & \na & \na & \na & 48.6$^{\dagger}$ \\
    video-SALMONN 2+ (3B) & Open, A+V & \na & \na & \na & \na & \na & \na & \na & \na & 48.3$^{\dagger}$ \\
    Gemini 1.5 Pro & Closed, A+V & 53.7 & 47.2 & 50.3 & 50.4 & 52.4 & 46.8 & 40.2 & 42.0 & 48.0 \\
    Qwen2.5-Omni (7B) & Open, A+V & 47.8 & 49.8 & 43.6 & 43.8 & 48.3 & 39.1 & 43.5 & 47.3 & 45.4 \\
    \ours \method (Qwen3-Omni-30B-A3B) & V+ASR+Audio & 38.8 & 42.7 & 37.1 & 38.3 & 39.3 & 34.3 & 36.7 & 37.4 & 38.0 \\
    VITA-1.5 (7B) & Open, A+V & 38.2 & 35.9 & 34.3 & 39.8 & 41.2 & 32.6 & 34.7 & 39.9 & 36.9 \\
    Unified-IO-2 XXL (7B) & Open, A+V & 27.1 & 31.7 & 23.9 & 23.7 & 25.5 & 23.7 & 25.7 & 27.3 & 25.9 \\
    VideoLLaMA2 (7B) & Open, A+V & 29.4 & 25.4 & 21.8 & 24.5 & 26.2 & 24.6 & 25.5 & 27.1 & 25.4 \\
    Unified-IO-2 XL (3B) & Open, A+V & 26.5 & 24.4 & 22.5 & 23.5 & 24.7 & 28.0 & 25.7 & 24.2 & 24.7 \\
    Unified-IO-2 L (1B) & Open, A+V & 19.3 & 22.8 & 23.1 & 25.6 & 25.8 & 24.1 & 22.9 & 25.3 & 23.3 \\
    OneLLM (7B) & Open, A+V & 26.7 & 25.1 & 19.0 & 22.7 & 27.0 & 23.7 & 22.4 & 19.8 & 22.8 \\
    \bottomrule
  \end{tabular}
\end{table*}

\begin{table*}[t]
  \caption{Full comparison on OmniVideoBench (1000 questions) by audio type and video duration in minutes (accuracy, \%).
  Baselines are the ``With Visual and Audio'' rows of the OmniVideoBench paper.
  Duration buckets follow its intervals, with the five videos just over 30\,min in (10,30].
  $^{\P}$ Overall score only, from the video-SALMONN-R3 paper.}
  \label{tab:appx_ovb}
  \centering
  \footnotesize
  \setlength{\tabcolsep}{4pt}
  \renewcommand{\arraystretch}{1.05}
  \begin{tabular}{llcccccccc}
    \toprule
    Model & Input & Music & Sound & Speech & (0,1] & (1,5] & (5,10] & (10,30] & Avg. \\
    \midrule
    \ours \method (Opus 5.5) & V+ASR+A2M & \best{59.34} & \best{69.39} & \best{72.18} & \best{67.84} & \second{73.37} & \best{71.62} & \second{67.94} & \best{70.60} \\
    \ours \method (Opus 5.5) & V+ASR & \second{57.14} & \second{68.03} & \second{71.78} & \second{64.91} & \best{74.85} & \second{67.69} & \best{68.70} & \second{69.90} \\
    Gemini-3.0-Pro & Closed, A+V & 52.81 & 55.17 & 64.13 & 62.42 & 66.18 & 57.02 & 59.76 & 61.80 \\
    \ours \method (Qwen3.8-27B) & V+ASR & 51.65 & 57.14 & 62.07 & 63.16 & 62.13 & 57.21 & 59.16 & 60.40 \\
    \ours \method (Qwen3.8-27B, medium) & V+ASR+Audio Caption & 47.25 & 56.46 & 62.07 & 61.99 & 61.54 & 55.02 & 60.69 & 59.90 \\
    Gemini-2.5-Pro & Closed, A+V & 38.46 & 57.72 & 61.66 & 57.83 & 64.43 & 55.02 & 55.94 & 58.90 \\
    \ours \method (Qwen3.8-27B, medium) & V+ASR & 45.05 & 55.10 & 60.50 & 59.06 & 59.17 & 58.95 & 56.11 & 58.30 \\
    Gemini-3.0-Flash & Closed, A+V & 49.45 & 50.34 & 56.69 & 58.43 & 55.10 & 55.90 & 52.29 & 55.10 \\
    Gemini-2.5-Flash & Closed, A+V & 39.56 & 57.04 & 53.17 & 55.42 & 55.10 & 47.37 & 52.11 & 52.40 \\
    video-SALMONN-R3 (8B) & Open, A+V & \na & \na & \na & \na & \na & \na & \na & 43.9$^{\P}$ \\
    Gemini-2.0-Flash & Closed, A+V & 29.67 & 40.27 & 43.21 & 49.40 & 43.15 & 41.05 & 34.87 & 41.50 \\
    Qwen3-Omni-30B-A3B & Open, A+V & 37.36 & 34.67 & 39.26 & 45.78 & 37.03 & 38.86 & 35.11 & 38.40 \\
    OmniVinci-9B & Open, A+V & 30.77 & 32.67 & 32.15 & 38.55 & 34.11 & 30.13 & 27.10 & 32.10 \\
    \ours \method (Qwen3-Omni-30B-A3B) & V+ASR+Audio & 30.77 & 27.21 & 32.55 & 43.86 & 27.81 & 30.57 & 29.39 & 31.60 \\
    Baichuan-Omni-1.5 & Open, A+V & 24.18 & 31.33 & 31.36 & 28.92 & 31.78 & 28.38 & 32.44 & 30.70 \\
    HumanOmni-7B & Open, A+V & 20.87 & 31.08 & 31.61 & 36.57 & 29.36 & 29.60 & 29.25 & 30.50 \\
    VITA-1.5-7B & Open, A+V & 25.27 & 28.57 & 31.49 & 31.33 & 27.41 & 30.57 & 33.97 & 30.50 \\
    MiniCPM-o & Open, A+V & 27.47 & 28.57 & 30.24 & 31.43 & 28.49 & 34.53 & 26.15 & 29.70 \\
    Qwen2.5-Omni-7B & Open, A+V & 23.07 & 25.33 & 30.70 & 41.57 & 27.41 & 25.33 & 26.72 & 29.30 \\
    VideoLLaMA2-7B & Open, A+V & 26.37 & 30.67 & 29.25 & 32.00 & 28.20 & 29.60 & 28.29 & 29.20 \\
    \bottomrule
  \end{tabular}
\end{table*}

\begin{table*}[t]
  \caption{Full comparison on LVOmniBench (1014 questions) by difficulty and ability (accuracy, \%).
  Baselines are the A+V rows of the LVOmniBench paper.
  We group our question types into its four abilities: Perception = Counting + Attribute Perception + Music Perception; Understanding = Human-Centric + Event Understanding; Inference = Sound + Spatial + Temporal Inference; Logical = Logical Reasoning.
  $^{\P}$ Overall score only, from the video-SALMONN-R3 paper.}
  \label{tab:appx_lvomni}
  \centering
  \footnotesize
  \setlength{\tabcolsep}{4pt}
  \renewcommand{\arraystretch}{1.05}
  \begin{tabular}{llcccccccc}
    \toprule
    Model & Input & Low & Medium & High & Underst. & Percept. & Infer. & Logical & Avg. \\
    \midrule
    \ours \method (Opus 5.5) & V+ASR+A2M & \second{81.2} & \best{77.1} & \best{68.7} & \second{80.0} & \best{74.9} & \best{72.7} & \best{82.6} & \best{76.2} \\
    \ours \method (Opus 5.5) & V+ASR & \best{82.5} & \second{76.0} & \second{66.4} & \best{80.4} & \second{74.4} & \second{70.6} & \best{82.6} & \second{75.5} \\
    Gemini-3.0-Pro & Closed, A+V & 79.3 & 68.1 & 45.0 & 73.3 & 60.1 & 65.4 & \second{67.5} & 65.8 \\
    \ours \method (Qwen3.8-27B, medium) & V+ASR+Audio Caption & 70.4 & 60.3 & 54.4 & 67.4 & 61.7 & 58.5 & 60.6 & 61.9 \\
    \ours \method (Qwen3.8-27B) & V+ASR & 68.8 & 62.6 & 51.7 & 66.5 & 59.8 & 60.6 & 61.5 & 61.7 \\
    \ours \method (Qwen3.8-27B, medium) & V+ASR & 69.4 & 61.7 & 46.3 & 62.6 & 58.5 & 59.9 & 61.5 & 60.2 \\
    Gemini-3.0-Flash & Closed, A+V & 76.6 & 63.0 & 31.0 & 67.7 & 54.7 & 60.6 & 51.8 & 59.0 \\
    Gemini-2.0-Flash & Closed, A+V & 57.0 & 48.9 & 29.8 & 41.4 & 38.5 & 49.1 & 42.1 & 42.9 \\
    video-SALMONN-R3 (8B) & Open, A+V & \na & \na & \na & \na & \na & \na & \na & 42.9$^{\P}$ \\
    Qwen3-Omni-30B & Open, A+V & 41.0 & 36.3 & 28.6 & 33.0 & 35.8 & 40.5 & 29.1 & 35.8 \\
    MiniCPM-o 4.5 & Open, A+V & 43.4 & 34.1 & 25.1 & 35.7 & 31.9 & 39.1 & 32.7 & 34.8 \\
    Ming-Omni-2.0-100B & Open, A+V & 41.3 & 32.9 & 29.3 & 30.0 & 36.5 & 33.9 & 39.1 & 34.6 \\
    video-SALMONN 2+ 7B & Open, A+V & 40.9 & 30.2 & 26.7 & 30.0 & 36.3 & 30.3 & 31.8 & 32.7 \\
    Qwen2.5-Omni-7B & Open, A+V & 37.7 & 29.9 & 28.3 & 29.1 & 34.9 & 31.8 & 28.2 & 32.0 \\
    \ours \method (Qwen3-Omni-30B-A3B) & V+ASR+Audio & 33.4 & 24.9 & 26.3 & 30.0 & 27.5 & 29.4 & 21.1 & 27.9 \\
    VideoLLaMA2-7B & Open, A+V & 27.0 & 26.8 & 28.2 & 23.9 & 28.2 & 29.4 & 24.6 & 27.2 \\
    \bottomrule
  \end{tabular}
\end{table*}

%-------------------------------------------------------------------------
\section{Reasoning models by question type}
\label{sec:appx_backbones}
\cref{tab:backbones} runs the harness on LVOmniBench with two omni reasoning models, which can also listen to the audio through \cmd{listen\_audio}.
\cref{tab:appx_backbones} breaks these runs down by the nine question types of LVOmniBench, and \cref{tab:appx_cost} gives their turns and time.
\method (Opus 5.5) reads only the transcript and is the most accurate overall and the best on seven of the nine types, with its largest lead, 5.5 points, on logical reasoning.
It is also the fastest: it answers in 9.5 turns and 5.3 minutes per question, while \method (Gemini 3.8 Flash) takes 2.8 times the turns and 3.4 times the time.
The harness raises the omni models as well, by 3.7 points for Gemini 3.8 Flash and 7.3 points for Qwen3.8-Omni-Flash over their results without it.
\cref{sec:appx_cases_agentic} follows one question of this comparison turn by turn.

\begin{table*}[t]
  \caption{The same harness with other reasoning models on LVOmniBench, by question type (accuracy, \%); \cref{tab:backbones} groups the same runs by difficulty and ability.
  V+ASR+Audio: an omni reasoning model that can also call \cmd{listen\_audio}.
  A+V rows: closed-source omni models without our harness, with their reported overall accuracy.
  \best{Bold}: best; \second{underline}: second best. Question counts in parentheses.}
  \label{tab:appx_backbones}
  \centering
  \footnotesize
  \setlength{\tabcolsep}{3.4pt}
  \renewcommand{\arraystretch}{1.1}
  \begin{tabular}{llcccccccccc}
    \toprule
    Model & Input & Attr. & Count. & Spat. & Event & Logic. & Human & Temp. & Music & Sound & Overall \\
          &       & \scriptsize(154) & \scriptsize(143) & \scriptsize(129) & \scriptsize(127) & \scriptsize(109) & \scriptsize(103) & \scriptsize(79) & \scriptsize(89) & \scriptsize(81) & \scriptsize(1014) \\
    \midrule
    \method (Opus 5.5)           & V+ASR       & \best{80.5} & \best{71.3} & \best{76.0} & \best{81.9} & \best{82.6} & \best{78.6} & \best{74.7} & \second{68.5} & 58.0 & \best{75.5} \\
    \method (Qwen3.8-Omni-Flash) & V+ASR+Audio & \second{73.4} & 65.7 & \second{74.4} & \second{78.0} & 67.9 & 68.9 & 72.2 & 65.2 & \second{66.7} & 70.6 \\
    \method (Gemini 3.8 Flash)   & V+ASR+Audio & \best{80.5} & \second{69.2} & 70.5 & \second{78.0} & \second{77.1} & \second{76.7} & \second{73.4} & \best{73.0} & \best{67.9} & \second{74.4} \\
    \midrule
    Qwen3.8-Omni-Flash           & A+V & \na & \na & \na & \na & \na & \na & \na & \na & \na & 63.3 \\
    Gemini 3.8 Flash             & A+V & \na & \na & \na & \na & \na & \na & \na & \na & \na & 70.7 \\
    Qwen3.5-Omni-Plus            & A+V & \na & \na & \na & \na & \na & \na & \na & \na & \na & 53.2 \\
    \bottomrule
  \end{tabular}
\end{table*}

\begin{table}[t]
  \caption{Turns and time per question of the runs of \cref{tab:appx_backbones} on LVOmniBench (accuracy in \%; time in minutes, without rate-limit waits).
  \method (Opus 5.5) is the most accurate and the fastest.}
  \label{tab:appx_cost}
  \centering
  \small
  \setlength{\tabcolsep}{4pt}
  \renewcommand{\arraystretch}{1.1}
  \begin{tabular}{lccc}
    \toprule
    Model & Acc. & Turns & Time \\
    \midrule
    \method (Opus 5.5)           & \best{75.5}   & \best{9.5}    & \best{5.3} \\
    \method (Gemini 3.8 Flash)   & \second{74.4} & 26.2          & 17.8 \\
    \method (Qwen3.8-Omni-Flash) & 70.6          & \second{24.6} & \second{14.5} \\
    \bottomrule
  \end{tabular}
\end{table}

%-------------------------------------------------------------------------
\section{The audio caption tool on music questions}
\label{sec:appx_audiocap}
\paragraph{Implementation.}
Audio Caption uses Ming-3.0-audio-tiny (9.3B parameters, 2.7B active).
A call to $\cmd{audio\_caption}(t_0,t_1,p)$ passes the original audio over $[t_0,t_1]$ and a question $p$ chosen by the VLM to this model, and returns its textual response.
The VLM uses the description to reason about musical tempo, mood, instruments and environmental sounds, and checks it against the frames and transcript of the same window.

\paragraph{Music results.}
\cref{tab:appx_audiocap} evaluates it on the 448 questions about the music itself: the questions a benchmark labels as music tasks, and those whose text asks about music, melody, rhythm or instruments rather than lyrics.
The model calls the tool on 84\% of them, and accuracy rises from 48.9\% to 58.7\%, by up to 18.7 points on WorldSense.
Offered on every question, the tool raises \method (Qwen3.8-27B, medium) on all four benchmarks: Daily-Omni from 80.3 to 81.3, WorldSense from 62.8 to 64.4, OmniVideoBench from 58.3 to 59.9 and LVOmniBench from 60.2 to 61.9 (rows V+ASR+Audio Caption in \cref{tab:appx_daily,tab:appx_worldsense,tab:appx_ovb,tab:appx_lvomni}).
The descriptions supply acoustic evidence that the speech transcript omits.
\cref{sec:appx_cases_caption} shows four of these calls step by step.

On the benchmarks' Music categories, Audio Caption raises accuracy from 52.2\% to 63.1\% on WorldSense (+10.9 points) and from 45.05\% to 47.25\% on OmniVideoBench (+2.2 points), with Qwen3.8-27B at medium effort (\cref{tab:appx_worldsense,tab:appx_ovb}).
These category scores use each benchmark's official grouping; \cref{tab:appx_audiocap} below evaluates the separate subset selected for questions about music itself.

\begin{table}[t]
  \caption{The \cmd{audio\_caption} tool raises \method (Qwen3.8-27B, medium) by 9.8 points on the 448 music questions (accuracy, \%).}
  \label{tab:appx_audiocap}
  \centering
  \small
  \setlength{\tabcolsep}{5pt}
  \renewcommand{\arraystretch}{1.1}
  \begin{tabular}{lcccc}
    \toprule
    & & \multicolumn{2}{c}{audio caption} & \\
    \cmidrule(lr){3-4}
    Benchmark & Questions & w/o & w/ & $\Delta$ \\
    \midrule
    Daily-Omni     & 34  & 64.7 & 70.6 & +5.9 \\
    WorldSense     & 193 & 43.5 & 62.2 & +18.7 \\
    OmniVideoBench & 85  & 51.8 & 51.8 & 0.0 \\
    LVOmniBench    & 136 & 50.7 & 55.1 & +4.4 \\
    \midrule
    Total          & 448 & 48.9 & 58.7 & +9.8 \\
    \bottomrule
  \end{tabular}
\end{table}

%-------------------------------------------------------------------------
\FloatBarrier
\section{Video-MME-v2}
\label{sec:appx_vmme2}
\cref{tab:five} adds Video-MME-v2 as a fifth benchmark, and \cref{tab:appx_vmme2} lists its full leaderboard with subtitles, which \method reads as its timed transcript.
\method (Opus 5.5) is the best model on every metric: its accuracy of 77.8\% is 6.8 points above Gemini 3.8 Flash and 5.1 above the Qwen Code agent of Gemini 3.8 Flash, and its non-linear score of 63.5 is 14.1 points above Gemini-3-Pro.
With Qwen3.8-27B, \method reaches 62.0\% accuracy, ahead of Gemini-3-Flash, GPT-5 and every open-source model on the leaderboard.

\begin{table*}[tp]
  \caption{Video-MME-v2 leaderboard with subtitles (w.\ sub; for omni models, with audio input), sorted by average accuracy.
  \method reads the official subtitles as its timed transcript.
  Non-Lin: grouped non-linear score; Level 1--3: the three levels of the benchmark; Cons.: capability consistency; Coh.: reasoning coherence; Acc.: average accuracy.
  Models evaluated with several frame counts are listed with their best setting.
  $^{\dagger}$: overall accuracy reported in the Qwen3.8-Omni technical report; Qwen-code: the model run as an agent in the Qwen Code harness.
  \best{Bold}: best; \second{underline}: second best; \na: not reported.}
  \label{tab:appx_vmme2}
  \centering
  \footnotesize
  \setlength{\tabcolsep}{6pt}
  \renewcommand{\arraystretch}{1.02}
  \begin{tabular}{lccccccc}
    \toprule
    Model & Non-Lin & Level 1 & Level 2 & Level 3 & Cons. & Coh. & Acc. \\
    \midrule
    \textit{Human Expert} & 90.7 & 94.8 & 91.1 & 87.9 & 91.7 & 88.9 & 94.9 \\
    \midrule
    \ours \method (Opus 5.5) & \best{63.5} & \best{79.0} & \best{66.2} & \best{52.6} & \best{64.5} & \best{61.7} & \best{77.8} \\
    Gemini 3.8 Flash (Qwen-code)$^{\dagger}$ & \na & \na & \na & \na & \na & \na & \second{72.7} \\
    Qwen3.8-Omni-Flash (Qwen-code)$^{\dagger}$ & \na & \na & \na & \na & \na & \na & 71.3 \\
    Gemini 3.8 Flash$^{\dagger}$ & \na & \na & \na & \na & \na & \na & 71.0 \\
    Gemini-3-Pro & \second{49.4} & \second{64.0} & \second{50.0} & \second{40.6} & \second{50.8} & \second{47.0} & 66.1 \\
    Qwen3.8-Omni-Flash$^{\dagger}$ & \na & \na & \na & \na & \na & \na & 65.0 \\
    Seed 2.0 Lite$^{\dagger}$ & \na & \na & \na & \na & \na & \na & 64.9 \\
    \ours \method (Qwen3.8-27B) & 44.4 & 59.5 & 47.7 & 33.2 & 45.5 & 42.4 & 62.0 \\
    Gemini-3-Flash & 42.5 & 58.3 & 44.8 & 31.7 & 44.9 & 38.2 & 61.1 \\
    Doubao-Seed-2.0-Pro-260215 & 43.3 & 54.4 & 47.0 & 34.1 & 43.7 & 42.5 & 60.5 \\
    Kimi-K2.5 & 40.7 & 51.8 & 43.4 & 32.2 & 40.6 & 40.9 & 58.0 \\
    MiMo-v2-Omni & 38.6 & 52.6 & 43.1 & 27.4 & 39.7 & 36.6 & 56.1 \\
    Qwen3.5-397B-A17B-Think & 39.1 & 50.3 & 41.8 & 30.7 & 39.0 & 39.4 & 55.9 \\
    GPT-5 & 37.0 & 44.5 & 39.1 & 31.1 & 37.4 & 36.3 & 55.6 \\
    Qwen3.5-27B-Think & 31.4 & 34.4 & 36.4 & 26.2 & 31.5 & 31.3 & 49.6 \\
    Qwen3.5-Omni-Plus$^{\dagger}$ & \na & \na & \na & \na & \na & \na & 47.9 \\
    Qwen3-VL-235B-A22B-Think & 28.1 & 32.6 & 30.3 & 23.9 & 28.3 & 27.8 & 47.2 \\
    Qwen3.5-Omni-Plus & 27.4 & 37.4 & 27.9 & 21.2 & 27.6 & 27.0 & 45.8 \\
    Qwen3.6-A3B-Think & 22.3 & 26.2 & 22.4 & 20.0 & 21.4 & 24.0 & 45.8 \\
    Gemma-4-31B & 28.2 & 30.5 & 29.2 & 26.1 & 26.7 & 30.9 & 45.6 \\
    Qwen3.5-122B-A10B-Think & 26.7 & 30.1 & 27.9 & 23.9 & 26.4 & 27.3 & 45.1 \\
    Qwen3.5-9B-Think & 26.2 & 30.5 & 30.1 & 20.9 & 25.8 & 26.8 & 44.5 \\
    Qwen3.5-35B-A3B-Think & 25.8 & 31.3 & 24.9 & 23.2 & 24.9 & 27.5 & 43.9 \\
    Qwen3-VL-235B-A22B-Instruct & 25.0 & 30.7 & 25.2 & 21.6 & 25.0 & 25.0 & 43.3 \\
    Qwen3.5-397B-A17B-Instruct & 24.5 & 27.3 & 25.9 & 21.9 & 23.9 & 25.7 & 42.1 \\
    Qwen3.5-27B-Instruct & 23.9 & 28.9 & 23.2 & 21.5 & 23.4 & 24.8 & 41.9 \\
    InternVL3.5-241B-A28B-Instruct & 23.1 & 28.2 & 23.7 & 19.6 & 23.4 & 22.5 & 41.4 \\
    Qwen3-VL-30B-A3B-Think & 22.1 & 27.6 & 22.4 & 18.6 & 21.7 & 22.8 & 39.8 \\
    GLM4.5V-Think & 21.8 & 26.2 & 21.7 & 19.4 & 20.5 & 24.3 & 39.2 \\
    Qwen3.5-122B-A10B-Instruct & 20.9 & 24.9 & 21.8 & 17.9 & 20.1 & 22.4 & 39.2 \\
    Qwen2.5-VL-72B-Instruct & 22.1 & 25.8 & 21.2 & 20.6 & 20.9 & 24.3 & 38.9 \\
    Qwen3.5-4B-Think & 20.7 & 26.2 & 21.0 & 17.3 & 20.4 & 21.4 & 38.8 \\
    GLM4.6-Flash & 19.8 & 26.8 & 19.4 & 15.9 & 19.6 & 20.0 & 37.4 \\
    Qwen3-VL-8B-Think & 20.2 & 23.5 & 21.4 & 17.4 & 20.1 & 20.3 & 37.3 \\
    KimiVL-16B-A3B-Instruct & 19.0 & 23.9 & 17.6 & 17.1 & 19.4 & 18.2 & 37.3 \\
    Qwen3.5-35B-A3B-Instruct & 19.9 & 24.2 & 18.6 & 18.4 & 18.4 & 22.7 & 36.9 \\
    Qwen3-Omni-30B-A3B-Think & 19.5 & 24.7 & 17.5 & 17.9 & 18.6 & 21.2 & 36.5 \\
    MiMo-VL-7B-RL-2508 & 19.4 & 24.9 & 18.8 & 16.6 & 18.9 & 20.3 & 36.0 \\
    Qwen3-VL-8B-Instruct & 18.2 & 21.8 & 17.0 & 16.9 & 17.1 & 20.3 & 35.7 \\
    Qwen3.5-9B-Instruct & 18.3 & 20.7 & 17.6 & 17.4 & 16.7 & 21.2 & 35.5 \\
    Qwen3-VL-4B-Think & 18.2 & 23.3 & 16.6 & 16.3 & 17.9 & 18.6 & 35.3 \\
    LLaVA-Video-72B-Qwen2 & 17.2 & 21.8 & 14.8 & 16.3 & 17.1 & 17.5 & 34.4 \\
    InternVL3.5-30B-A3B-Instruct & 16.6 & 21.6 & 14.5 & 15.2 & 15.9 & 18.0 & 34.2 \\
    Qwen3-VL-4B-Instruct & 17.6 & 23.6 & 14.6 & 16.2 & 16.3 & 19.9 & 34.1 \\
    Qwen3-Omni-30B-A3B-Instruct & 17.1 & 21.7 & 15.0 & 16.0 & 16.5 & 18.3 & 34.1 \\
    Qwen3-VL-30B-A3B-Instruct & 16.8 & 18.0 & 17.6 & 15.6 & 15.7 & 19.0 & 33.2 \\
    InternVL3.5-8B-Instruct & 16.4 & 21.6 & 15.3 & 14.2 & 16.1 & 17.1 & 32.4 \\
    KimiVL-16B-A3B-Think & 15.7 & 21.8 & 14.0 & 13.3 & 14.9 & 17.1 & 32.4 \\
    Qwen3.5-4B-Instruct & 15.4 & 17.6 & 16.1 & 13.5 & 14.6 & 16.7 & 32.3 \\
    MiMo-VL-7B-SFT-2508 & 14.7 & 19.8 & 13.2 & 12.8 & 14.1 & 15.8 & 31.2 \\
    Keye-VL-1.5-8B & 14.1 & 19.1 & 12.4 & 12.4 & 13.7 & 14.8 & 31.1 \\
    Qwen2.5-VL-7B-Instruct & 14.4 & 20.0 & 13.0 & 12.1 & 13.7 & 15.6 & 30.9 \\
    VideoLLaMA3-7B & 14.2 & 19.0 & 12.7 & 12.4 & 14.3 & 13.9 & 30.6 \\
    InternVL3.5-4B-Instruct & 14.2 & 21.6 & 10.5 & 12.6 & 14.1 & 14.6 & 30.5 \\
    GLM4.1V-Think & 14.0 & 18.8 & 13.2 & 11.8 & 13.8 & 14.4 & 30.3 \\
    VITA-1.5-7B & 11.3 & 14.1 & 9.4 & 11.0 & 10.3 & 13.0 & 25.8 \\
    LLaVA-Video-7B-Qwen2 & 9.7 & 15.9 & 7.4 & 7.5 & 9.9 & 9.0 & 24.0 \\
    \bottomrule
  \end{tabular}
\end{table*}

\begin{figure*}[t]
  \centering
  \includegraphics[width=\textwidth]{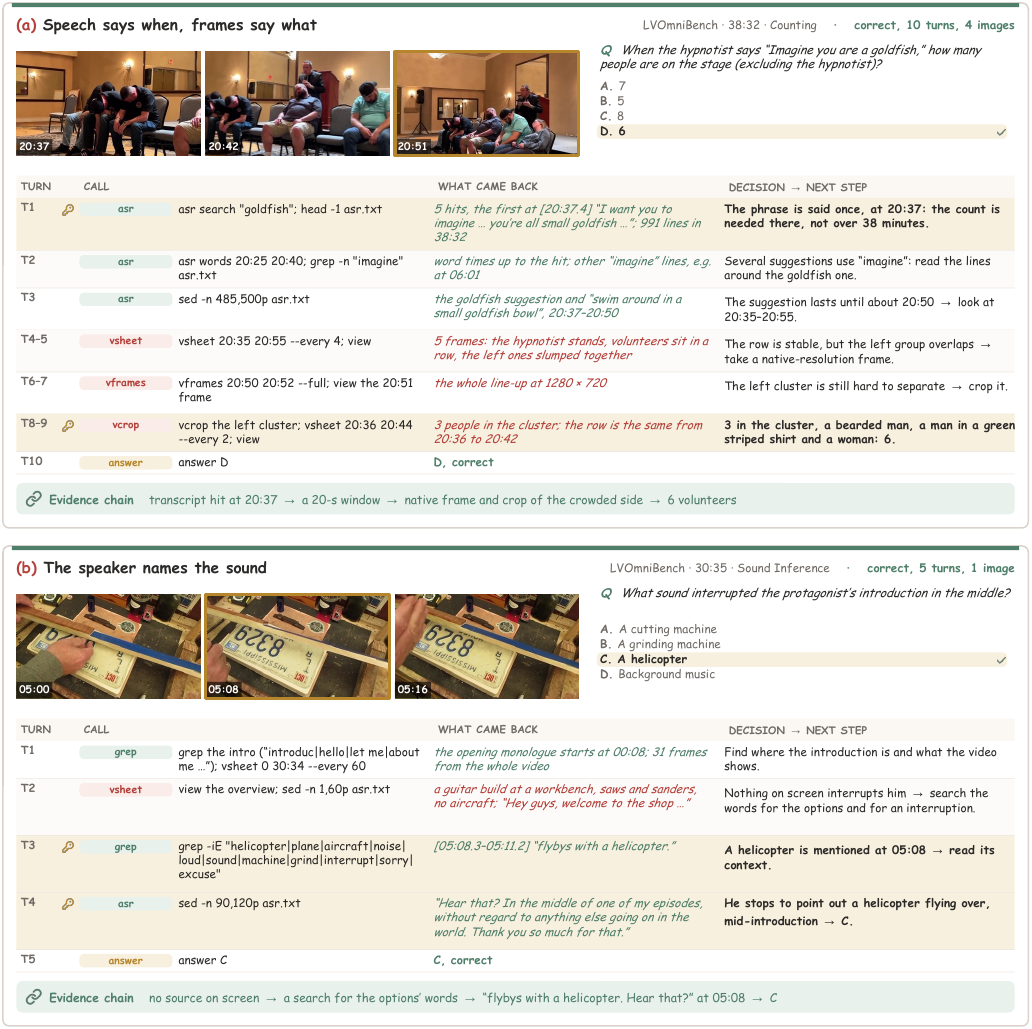}
  \caption{Step-by-step trajectories answered from the timed transcript. (a)~A count at the moment a phrase is said in a 38-minute video. (b)~A sound that never appears on screen and is named by the speaker. Panels (a) and (b) are \cref{fig:cases}a and~b.}
  \label{fig:appx_cases1}
\end{figure*}

\begin{figure*}[t]
  \centering
  \includegraphics[width=\textwidth]{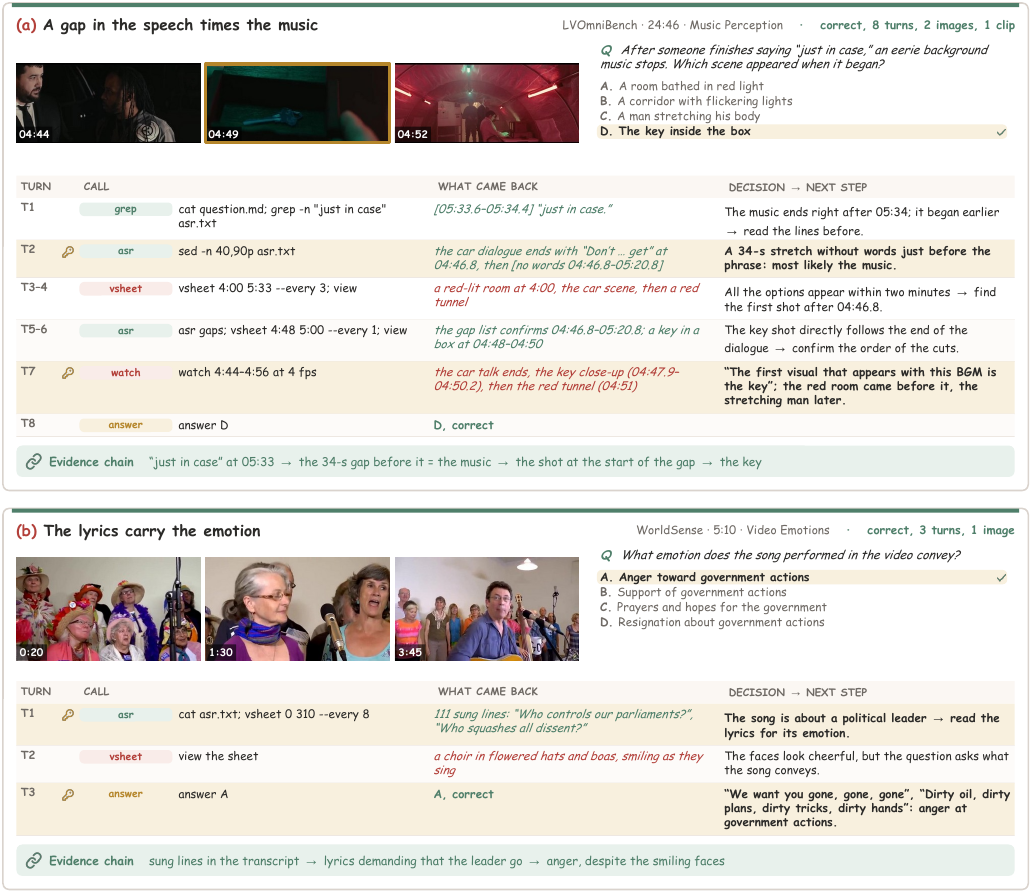}
  \caption{Step-by-step trajectories answered from the timed transcript. (a)~A gap in the speech that times the music. (b)~Lyrics that carry the emotion of a song.}
  \label{fig:appx_cases2}
\end{figure*}

\begin{figure*}[t]
  \centering
  \includegraphics[width=\textwidth]{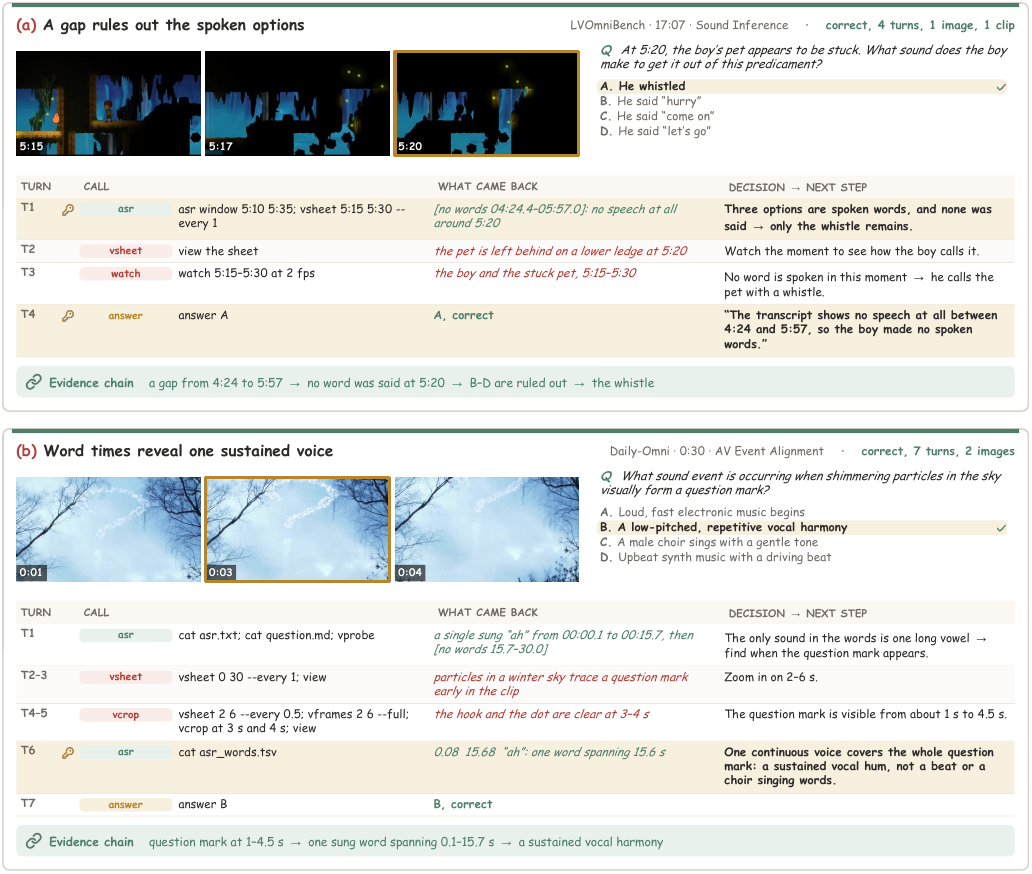}
  \caption{Step-by-step trajectories answered from the timed transcript. (a)~A gap that rules out the spoken options. (b)~Word times that reveal one sustained voice.}
  \label{fig:appx_cases3}
\end{figure*}

\begin{figure*}[t]
  \centering
  \includegraphics[width=\textwidth]{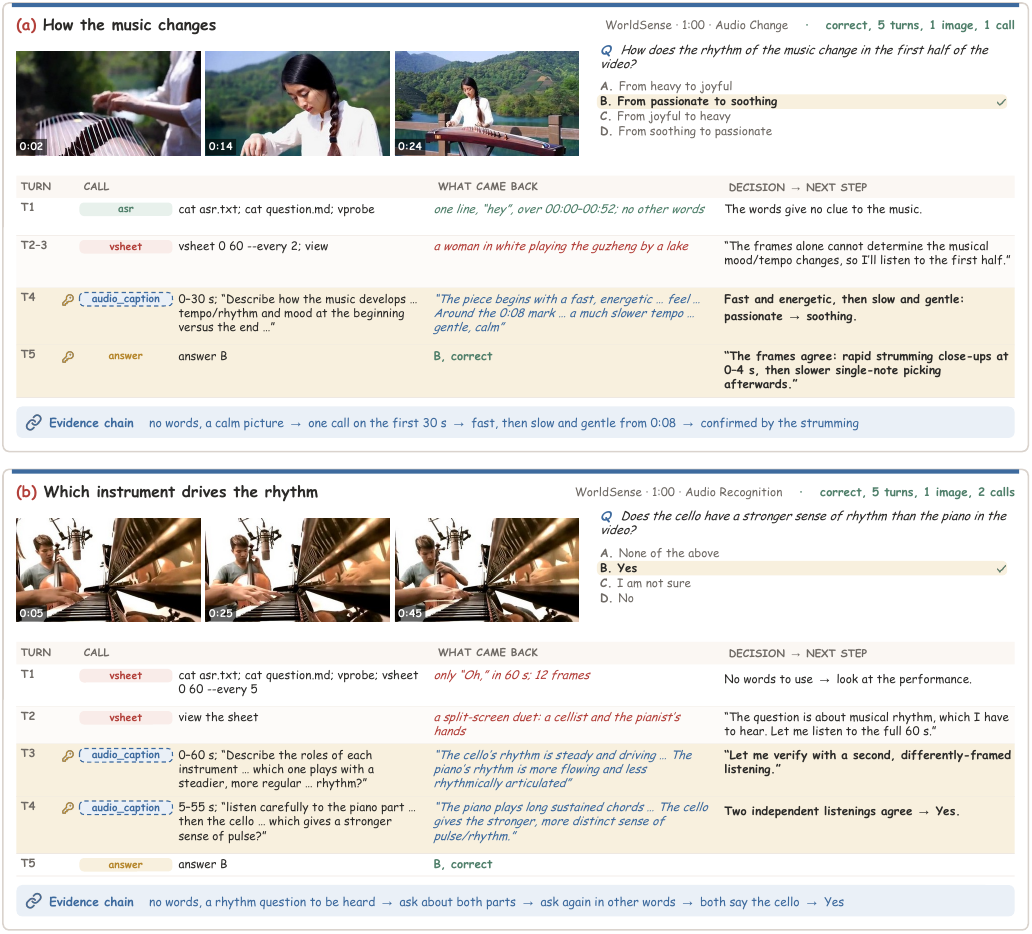}
  \caption{Step-by-step trajectories with the \cmd{audio\_caption} tool. (a)~How the music changes, \cref{fig:cases}c. (b)~Which instrument drives the rhythm, settled by two differently worded calls.}
  \label{fig:appx_cases4}
\end{figure*}

\begin{figure*}[t]
  \centering
  \includegraphics[width=\textwidth]{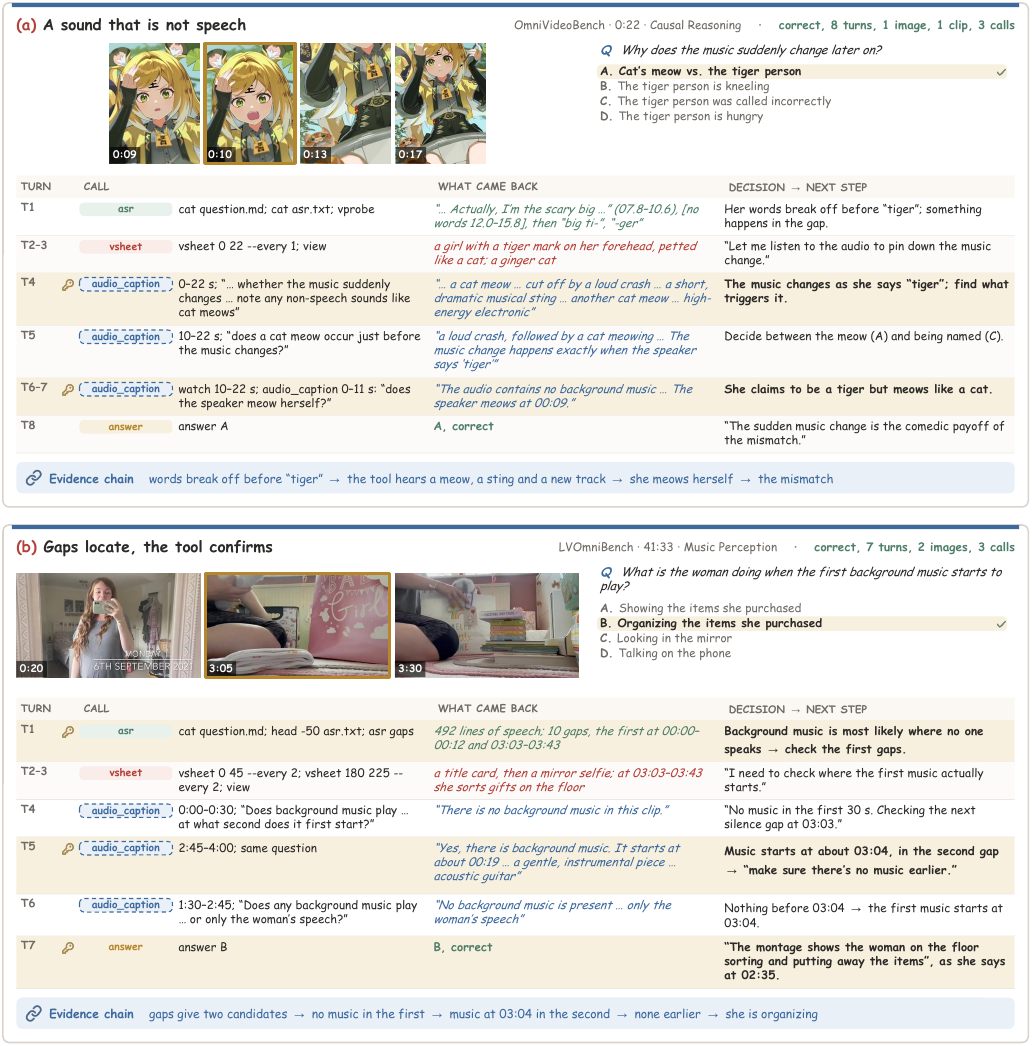}
  \caption{Step-by-step trajectories with the \cmd{audio\_caption} tool. (a)~A non-speech sound that explains the plot. (b)~The start of the first music in a 41-minute video, located by the transcript gaps and confirmed by the tool.}
  \label{fig:appx_cases5}
\end{figure*}

\begin{figure*}[t]
  \centering
  \includegraphics[width=\textwidth]{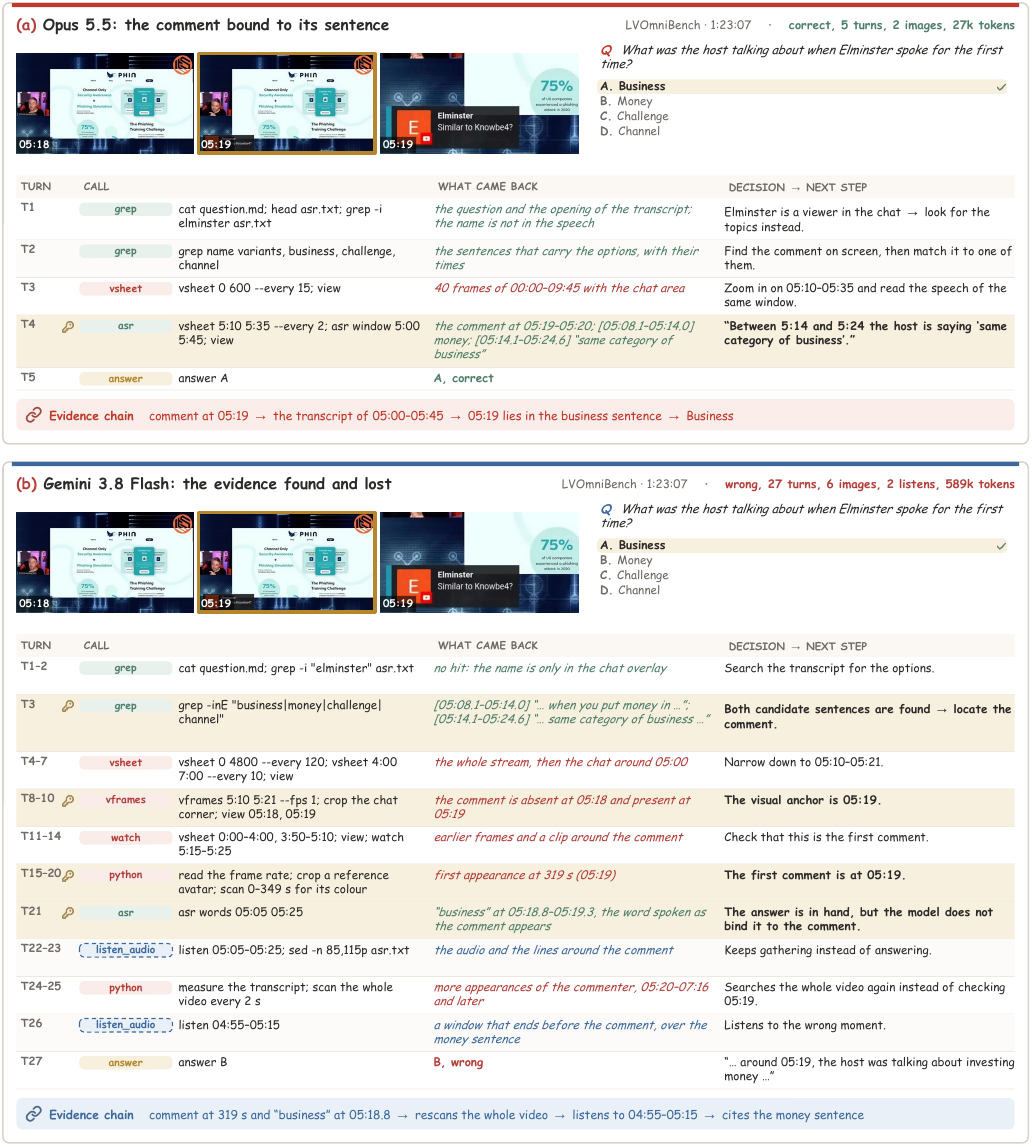}
  \caption{The same question in the same harness with two reasoning models, \cref{fig:cases}d. (a)~\method (Opus 5.5), a VLM, binds the comment to the sentence spoken at that moment in five turns. (b)~\method (Gemini 3.8 Flash), an omni model, finds the same evidence and keeps searching for 27 turns.}
  \label{fig:appx_cases6}
\end{figure*}

\FloatBarrier
%-------------------------------------------------------------------------
\section{Case studies in detail}
\label{sec:appx_cases}
\cref{sec:cases} shows four trajectories with the decisive steps highlighted.
This section gives ten trajectories of \method (Qwen3.8-27B, medium) and the comparison of \cref{fig:cases}d step by step, with every call, what came back and the decision it led to.
Decisions in quotation marks are the notes the model wrote between its calls; the others state what its next call shows it concluded.
Highlighted rows supplied the decisive evidence.
The three subsections follow \cref{sec:cases}: answering from the timed transcript, hearing the music through a tool, and a VLM and an omni model in one harness.

\subsection{Answering from the timed transcript}
\label{sec:appx_cases_asr}
In the six cases of \cref{fig:appx_cases1,fig:appx_cases2,fig:appx_cases3}, the transcript does three jobs: it is an index that tells the model where to look, a record of what was said or sung, and, through its gaps and word times, a clock for the sounds that carry no words.

\paragraph{The transcript as a temporal index (\cref{fig:appx_cases1}a).}
The model starts from the words of the question rather than from the picture.
At T1, one search for ``goldfish'' returns the suggestion at 20:37 in a transcript of 991 lines, so the count is needed only there.
At T2--T3, since other suggestions also begin with ``imagine'', it reads the word times and the surrounding lines, which bound this one to 20:37--20:50.
At T4--T5, a contact sheet of 20:35--20:55 shows the volunteers seated in a row, unchanged over the window, with those on the left slumped against each other.
At T6--T9, it extracts a native-resolution frame at 20:51, crops the left cluster and checks with a denser sheet that the row stays the same: three people in the cluster and three more.
It answers six after viewing four images, all within 20 seconds of a 38-minute video.

\paragraph{A sound named in speech (\cref{fig:appx_cases1}b).}
At T1--T2, the model locates the introduction and takes an overview of the 30-minute video: a guitar build at a workbench, with saws and sanders but no aircraft in view, so the picture does not show what interrupted the speaker.
At T3, it searches the transcript for the words of all four options and for signs of an interruption, and the one matching option word is ``flybys with a helicopter'' at 05:08.
At T4, the next lines, ``Hear that? In the middle of one of my episodes \ldots{} Thank you so much for that'', show the speaker stopping to point out a helicopter flying over, and the model answers after five turns and a single image.

\paragraph{A gap that times the music (\cref{fig:appx_cases2}a).}
At T1, ``just in case'' is found at 05:33.6, so the music ends right after it and began earlier.
At T2, the lines before it show the car dialogue ending at 04:46.8 and a 34-second gap before speech resumes at 05:20.8; the model reads this stretch without words as the music.
At T3--T4, a sheet of 4:00--5:33 shows that the scenes of all four options occur within two minutes, so the order of the shots decides.
At T5--T6, the gap list confirms the stretch, and a dense sheet finds a key in a box at 04:48--04:50, right after the dialogue ends.
At T7, a clip of 4:44--4:56 fixes the order of the cuts, the car, then the key from 04:47.9, then the red tunnel at 04:51, and the model concludes that the key is ``the first visual that appears with this BGM''.

\paragraph{Lyrics (\cref{fig:appx_cases2}b).}
At T1, the transcript holds 111 sung lines, such as ``Who controls our parliaments?'' and ``Who squashes all dissent?'', and the model reads the message of the song from them.
At T2, the sheet shows a choir in flowered hats smiling as they sing; the model keeps to what the question asks, the emotion the song conveys.
At T3, lines such as ``We want you gone, gone, gone'' decide it: anger toward government actions, in three turns.

\paragraph{A gap that rules out the spoken options (\cref{fig:appx_cases3}a).}
At T1, the model prints the transcript around 5:20 and finds one gap from 4:24 to 5:57: no word is said at that moment, which rules out the three options that are spoken words.
At T2--T3, a sheet and a clip of 5:15--5:30 show the pet stuck on a lower ledge and the boy with it.
At T4, the model answers that he whistled, because ``the transcript shows no speech at all between 4:24 and 5:57''.

\paragraph{Word times that reveal a sustained voice (\cref{fig:appx_cases3}b).}
At T1, the transcript of this 30-second clip holds a single sung ``ah'' from 0.1 to 15.7\,s.
At T2--T5, a sheet, a denser sheet, native-resolution frames and two crops place the question mark at about 1--4.5\,s.
At T6, the word file shows that the ``ah'' is one word lasting 15.6\,s, so one continuous voice spans the whole question mark: a sustained vocal hum, not music with a beat or a choir singing words.

\subsection{Hearing the music through a tool}
\label{sec:appx_cases_caption}
When the question is about the sound itself, the model calls \cmd{audio\_caption} with a question of its own and checks the reply against the frames and the transcript.
In \cref{fig:appx_cases4,fig:appx_cases5}, it asks how the music develops, rephrases a question to confirm a reply, narrows the window to time an event, and lets the transcript gaps decide where to listen.

\paragraph{How the music changes (\cref{fig:appx_cases4}a).}
At T1, the transcript holds one word over 52\,s, so the words give no clue.
At T2--T3, the sheet shows a guzheng player by a lake, and the model notes that ``the frames alone cannot determine the musical mood/tempo changes'' and decides to listen.
At T4, one call on the first half, asking how tempo and mood develop, returns a fast, energetic opening that turns slow and gentle near 0:08, that is, from passionate to soothing.
At T5, the model checks the reply against the frames, ``rapid strumming close-ups at 0--4\,s, then slower single-note picking'', before answering.

\paragraph{Two calls that agree (\cref{fig:appx_cases4}b).}
At T1--T2, with only ``Oh,'' in the transcript and a duet of cello and piano on screen, the model recognizes that rhythm is something ``I have to hear''.
At T3, it asks the tool to describe the role and rhythm of each part over the full minute: the cello is ``steady and driving'', the piano ``more flowing''.
At T4, it asks again over 5--55\,s in other words, listening to each part in turn, and the second reply agrees that the cello gives the stronger pulse; the model answers yes.

\paragraph{A non-speech sound (\cref{fig:appx_cases5}a).}
At T1, the girl's words break off before ``tiger'', with a gap at 12.0--15.8\,s.
At T2--T3, the frames show a girl with a tiger mark on her forehead, petted like a cat.
At T4, a first call over the whole clip reports a cat's meow, a crash, a musical sting and a switch to electronic music.
At T5, a second call on 10--22\,s places the change exactly at the word ``tiger'', with a meow right before it, which separates the meow (A) from being named (C).
At T6--T7, after watching the moment, the model asks whether the speaker meows herself; she does, at 0:09.
A self-declared tiger who meows is the joke, and the model answers A.

\paragraph{Gaps and the tool together (\cref{fig:appx_cases5}b).}
At T1, in a 41-minute vlog with 492 lines of speech, the model lists the gaps, where background music is most likely; the first two are at 00:00--00:12 and 03:03--03:43.
At T2--T3, sheets of both show a title card and a mirror selfie, then the woman sorting gifts on the floor.
At T4, the tool hears no music in the first 30\,s, so the model moves to the next gap.
At T5, over 2:45--4:00, the music starts about 19\,s in, at 03:04, inside the second gap.
At T6, to make sure that this is the first music, the model checks the speech before it, 1:30--2:45, and finds none.
At T7, the frames at 03:04 show the woman sorting and putting away the gifts, as she announces at 02:35, and the model answers after three short calls.

\subsection{A VLM and an omni model in one harness}
\label{sec:appx_cases_agentic}
\cref{sec:cases:agentic} compares two reasoning models on one LVOmniBench question; \cref{fig:appx_cases6,tab:appx_compare} give it turn by turn.

\begin{table}[t]
  \caption{The question of \cref{fig:appx_cases6} in the same harness: the VLM answers correctly with a fraction of the turns, tokens and time of the omni model.}
  \label{tab:appx_compare}
  \centering
  \small
  \setlength{\tabcolsep}{3pt}
  \renewcommand{\arraystretch}{1.1}
  \begin{tabular}{lrr}
    \toprule
    & Opus 5.5 & Gemini 3.8 Flash \\
    \midrule
    Answer                      & \best{A (correct)} & B (wrong) \\
    Turns                       & \best{5}       & 27 \\
    Tool calls (w/o answer)     & \best{6}       & 26 \\
    Images viewed               & \best{2}       & 6 \\
    Audio listens               & 0              & 2 \\
    Tokens                      & \best{27{,}378} & 589{,}458 \\
    Time per question (s)       & \best{75.8}    & 588.7 \\
    \bottomrule
  \end{tabular}
\end{table}

\paragraph{The question and the evidence.}
The question comes from an 83-minute live stream and asks what the host was talking about when a viewer named Elminster commented for the first time.
Both models work in the same harness, with the same transcript and the same tools, and Gemini 3.8 Flash can also listen to the original audio.
The question is decided by a single check in time.
The comment first appears at 319\,s (05:19).
The host says ``when you put money in, you like to know how that money's doing'' over 308.1--314.0\,s and then ``similar to No Before, same category of business'' over 314.1--324.6\,s, with the word ``business'' at 318.8--319.3\,s.
Since 319\,s lies in the second sentence, the answer is Business.

\paragraph{The VLM binds the evidence in five turns.}
At T1, \method (Opus 5.5) reads the question and the transcript and sees that the name is not in the speech: Elminster is a viewer whose comment is shown on screen.
At T2, it searches the transcript for the words of the options and obtains the candidate sentences with their times.
At T3, one contact sheet of the first ten minutes shows where the chat appears.
At T4, it zooms in on 05:10--05:35 and, in the same turn, reads the transcript of 05:00--05:45, so that the picture and the speech of one window are in front of it together.
The comment appears at 05:19--05:20, and the model binds it to the sentence spoken at that moment: ``Between 5:14 and 5:24 the host is saying `same category of business'\,''.
It answers Business at T5, after two images and 27k tokens (\cref{tab:appx_compare}).
Every call serves the one open question, where the comment falls in the speech, and the search stops as soon as that question is settled.

\paragraph{The omni model finds the evidence and keeps searching.}
\method (Gemini 3.8 Flash) collects every piece of this evidence.
At T3, its search of the transcript returns both the money sentence and the business sentence.
At T8--T10, crops of single frames show the comment absent at 05:18 and present at 05:19.
At T15--T20, it crops a reference avatar and scans the frames for its colour, which returns 319\,s as the first appearance.
At T21, the word times put ``business'' at 05:18.8--05:19.3, the very moment of the comment.
At this point the moment, the sentence and the word are all in its history.
It does not use them.
At T22--T23, it listens to 05:05--05:25 and rereads the transcript of the same lines; at T24--T25, it scans the whole video for later appearances of the commenter, which adds appearances from 05:20 to 07:16 and beyond but nothing about 05:19; and at T26, it listens to 04:55--05:15, a window that covers the money sentence and ends four seconds before the comment.
At T27, it answers Money with the reason ``when Elminster's chat message appeared on screen for the first time around 05:19, the host was talking about investing money''.
Its own reason places the comment at 05:19 and quotes a sentence that ended at 05:14.
The omni model spends 27 turns, 21 times the tokens and 7.8 times the time of the VLM, and answers Money where the VLM answers Business (\cref{tab:appx_compare}).

\paragraph{Where the two trajectories diverge.}
The two trajectories diverge at the three points an agent depends on.
First, it does not bind evidence across modalities: it holds the visual moment and the spoken word for that moment, but never joins them into one fact.
Second, its later calls do not converge on the open question: after the first appearance is settled, it rescans the whole video and listens to a window that does not contain the moment it has already found.
Third, it does not check its answer against its own evidence: a reason that names 05:19 and quotes a sentence that ended at 05:14 contradicts itself.
The VLM does all three within a single turn.
This question is decided by binding a moment on screen to the words spoken there, and both models already hold those words, in the transcript and, for the omni model, in the audio as well.
Planning, evidence binding and self-checking therefore decide it, and strengthening them is the most direct way to raise an omni model, which already hears the audio, on such questions.
Across LVOmniBench, the same pattern costs \method (Gemini 3.8 Flash) 26.2 turns per question against 9.5 for \method (Opus 5.5), at lower accuracy (\cref{tab:appx_cost}).

%-------------------------------------------------------------------------
\section{Audio to Math: signal analysis as agent tools}
\label{sec:appx_audiomath}

Audio to Math (A2M) makes music and environmental sounds accessible to the VLM through numerical measurements, tables and plots.
The extension adds \cmd{audio\_analyze} for standard signal analysis and \cmd{audio\_python} for analyses written by the model.
The model receives their numerical outputs as text, inspects saved tables and plots with the existing \cmd{bash} and \cmd{view} tools, and aligns them with frames, clips and word times on the original video timeline.

\begin{figure*}[t]
  \centering
  \includegraphics[width=\textwidth]{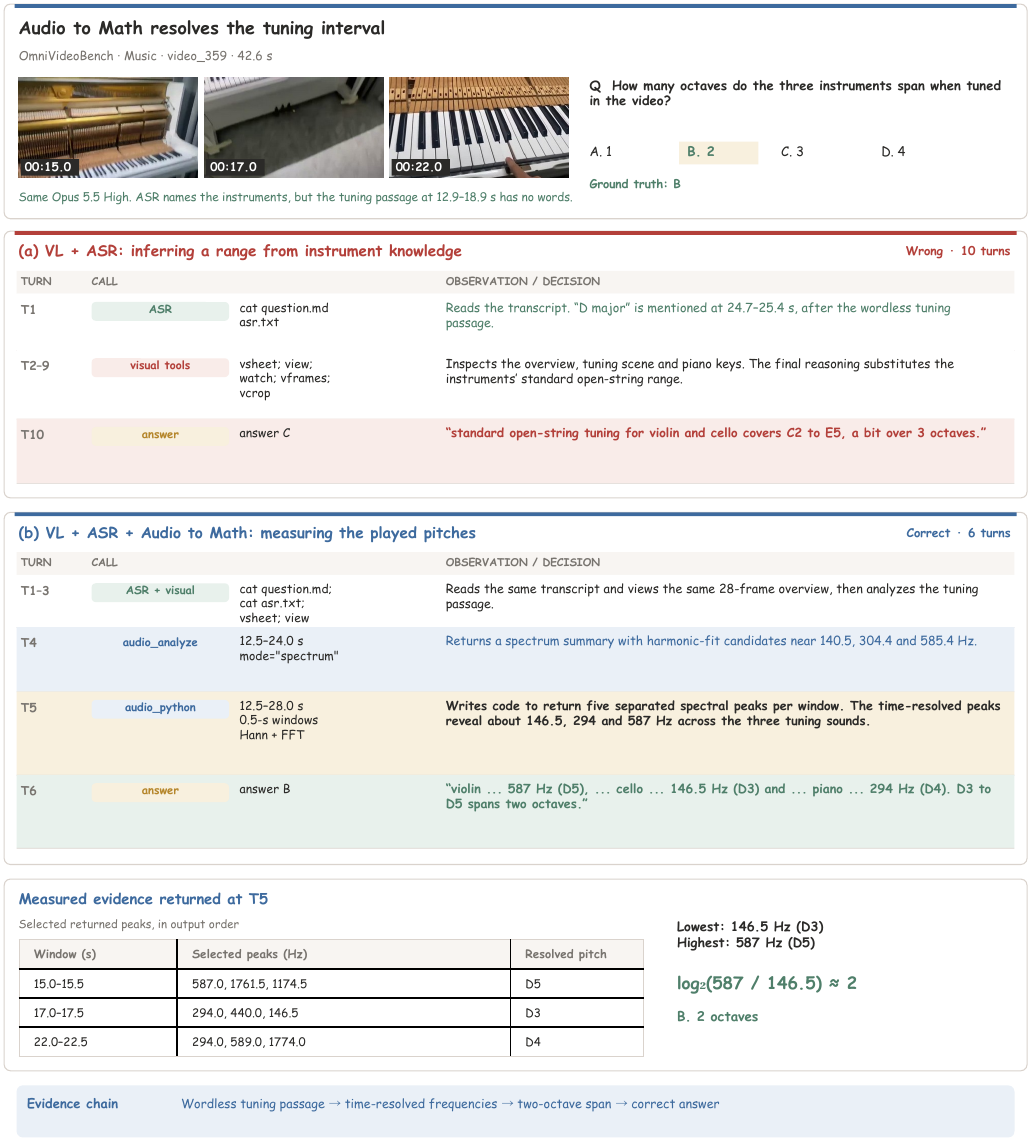}
  \caption{Audio to Math resolves the tuning interval in OmniVideoBench \cmd{video\_359}. Both trajectories use Opus 5.5 at high effort. (a)~Visual tools and ASR lead to an incorrect three-octave estimate after ten turns. (b)~\cmd{audio\_analyze} and model-written \cmd{audio\_python} measurements recover pitches near 146.5, 294 and 587\,Hz, yielding the correct two-octave answer in six turns.}
  \label{fig:appx_audiomath_case}
\end{figure*}

\subsection{Recovering the played pitches}

\cref{fig:appx_audiomath_case} compares two trajectories for the question ``How many octaves do the three instruments span when tuned in the video?''
The transcript names the instruments, but contains no words during the tuning passage at 12.9--18.9\,s.
With visual tools and ASR, Opus 5.5 inspects the overview, tuning scene and piano keys, then substitutes the instruments' standard open-string range for the pitches actually played.
It answers three octaves at T10, incorrectly.

With A2M, the same model first reads the transcript and views the contact sheet at T1--T3.
At T4, it calls \cmd{audio\_analyze} in spectrum mode on 12.5--24.0\,s to obtain harmonic-fit candidates.
At T5, it writes \cmd{audio\_python} code to resolve the peaks over time within 12.5--28.0\,s.
The returned frequency patterns support tuning pitches near 587\,Hz for the violin, 146.5\,Hz for the cello and 294\,Hz for the piano.
Their highest-to-lowest frequency ratio gives
\begin{equation}
  \Delta_{\mathrm{octave}}=\log_2\!\left(\frac{587}{146.5}\right)\approx 2.
\end{equation}
At T6, the model identifies D3, D4 and D5 and correctly answers two octaves.
The signal measurements recover the pitch registers omitted by ASR and settle the question with fewer turns than repeated visual inspection.

\subsection{Standard analysis with \cmd{audio\_analyze}}

The interface is $\cmd{audio\_analyze}(t_0,t_1,m,\theta)$, where $m$ selects an analysis mode and $\theta$ supplies its parameters.
Times are absolute video times, accepted as seconds or \cmd{mm:ss}.
The five modes provide the following measurements:
\begin{itemize}
  \item \textbf{Overview.} Energy, frequency structure, periodicity and candidate acoustic changes, together with a compact overview plot.
  \item \textbf{Events.} Transient candidates with onset and peak times, peak levels, detector thresholds and an event plot.
  \item \textbf{Spectrum.} Frequency peaks and time--frequency structure within the selected window.
  \item \textbf{Rhythm.} Periodicity candidates and the timing of detected events, for comparing repeated sounds and musical pulses.
  \item \textbf{Compare.} Recorded levels and spectral measurements for the selected window and a second window specified in \cmd{comparison\_window}.
\end{itemize}
Parameters include the frequency band, channel, analysis window and hop, minimum event spacing and detection thresholds.
For example, selecting a lower-frequency band lets the model examine bass pulses separately from higher-frequency transients.
The model chooses these settings from the question and the observed signal structure.

The standard analysis computes a short-time Fourier transform with a Hann window, framewise RMS level, spectral centroid, onset strength and spectral change.
Its default window is 42.7\,ms with a 10\,ms hop; the analysis records the actual window, hop, sample rate and selected band.
For a frame $y_n$ of $N$ normalized samples, its RMS level is
\begin{equation}
  L_{\mathrm{RMS}} = 20\log_{10}\!\left(\max\!\left[\epsilon,
  \sqrt{\frac{1}{N}\sum_{n=1}^{N}y_n^2}\right]\right),
\end{equation}
reported in dBFS, with a numerical floor $\epsilon$.
The detector records each candidate's onset, peak time, peak level, level above the local floor and acceptance under the selected thresholds.
The model derives counts from the accepted events and compares their timestamps with the corresponding visual events.
Periodicity estimates support this inspection; source identity and musical interpretation come from the combined signal, visual and transcript evidence.

\paragraph{Returned observations.}
Each response contains the analyzed window, timeline reference, channel, parameters, a compact numerical summary and paths relative to the sandbox.
The full outputs include \cmd{features.csv}, \cmd{events.csv}, \cmd{events.json}, \cmd{analysis.json} and PNG plots.
The feature table carries absolute times, RMS levels, spectral centroids, onset strengths and spectral changes; the event table retains both accepted and rejected candidates with their detector decisions.
The VLM receives the measurements as text, opens the returned PNGs with \cmd{view}, and uses \cmd{bash} to filter tables or compare measurements across windows.
This keeps the measurements available for later turns without repeating the analysis.

\subsection{Model-written analysis with \cmd{audio\_python}}

The interface is $\cmd{audio\_python}(t_0,t_1,\mathrm{code})$.
It executes Python written by the model on the selected original audio window.
The worker provides \cmd{x}, a float32 array of samples by channels; \cmd{sr}, the source sample rate; \cmd{t}, the absolute timestamp of every sample; the window bounds; and NumPy, SciPy and Matplotlib.
The model can design a band filter, compare amplitude envelopes, inspect harmonics or calculate timing differences for the event it is investigating.
It assigns a compact JSON-compatible finding to \cmd{result} and writes larger outputs with \cmd{save\_json}, \cmd{save\_csv} and \cmd{save\_plot}.

In \cref{fig:appx_audiomath_case}, the model averages the audio channels, divides the signal into 0.5-second windows, applies a Hann window and computes a four-times zero-padded FFT.
For each window, its code selects five peaks between 60 and 2000\,Hz, separated by more than 15\,Hz, and returns their frequencies with the absolute start time and the RMS level of the Hann-windowed signal.
The 17.0--17.5\,s window contains peaks at 294.0, 440.0 and 146.5\,Hz: the harmonic pattern supports a 146.5\,Hz fundamental even though 294.0\,Hz is the strongest peak.
The model combines these time-resolved measurements with the instrument context to determine the octave span.

The tool returns this result, printed output and paths to the generated artifacts.
The model can then open a plot with \cmd{view}, read a table with \cmd{bash}, refine the window or analysis code, and revisit the matching video interval with \cmd{vsheet}, \cmd{vframes}, \cmd{vcrop} or \cmd{watch}.
The analysis is chosen during the trajectory, so the same interface supports different musical and environmental sounds.

\subsection{Harness integration and evaluation}

The \cmd{audio-math-v1} option enables both tools.
The initial evidence-gathering stage uses the original visual and ASR operations; the model considers audio analysis during localization when the question still needs acoustic evidence.
On first use, the harness copies the original audio stream into \cmd{audio\_source.mka} in the sandbox and records its timing metadata in \cmd{audio\_source.json}.
The visual tools continue to use the muted video, while both analysis tools decode the selected interval from the original source.
Results carry the original video timestamps so that the model can compare measurements directly with word times and visible actions.

The extension allows twelve analysis calls per question, windows of up to 120\,s and 90\,s per CPU analysis worker, within the overall question budget.
Identical successful calls reuse their cached results.
The trace records each tool call, its parameters and code, returned observations and artifact paths, preserving the evidence used for the answer.

With Opus 5.5 at high effort, A2M raises Music accuracy from 64.8\% to 68.0\% on WorldSense (+3.2 points) and from 57.14\% to 59.34\% on OmniVideoBench (+2.2 points; \cref{tab:appx_worldsense,tab:appx_ovb}).
With Qwen3.8-27B at medium effort, Audio Caption raises the corresponding scores from 52.2\% to 63.1\% (+10.9 points) and from 45.05\% to 47.25\% (+2.2 points).
Each gain compares the same reasoning model with and without the respective tool family.
Both routes improve music understanding by supplying acoustic evidence that the transcript omits.

\input{sec/appendix_prompts}

%% file: sec/appendix_prompts.tex
\clearpage
\section{Core prompts for the three harness variants}
\label{sec:appx_prompts}

The blocks below collect the core system instructions and tool descriptions.
VL+ASR supplies the shared visual and transcript workflow; A2M and Audio Caption add the corresponding audio guidance.

\definecolor{promptbase}{HTML}{4F7F6A}
\definecolor{promptmath}{HTML}{3D6A9E}
\definecolor{promptaudio}{HTML}{B5862B}
\newcommand{\coreprompt}[3]{%
  \par\begingroup
  \setlength{\fboxsep}{9pt}%
  \setlength{\fboxrule}{0.5pt}%
  \noindent\fcolorbox{#1!65}{#1!3}{%
    \begin{minipage}{\dimexpr\linewidth-2\fboxsep-2\fboxrule\relax}
      \fontsize{9.2}{11.5}\selectfont
      \setlength{\parindent}{0pt}%
      \setlength{\parskip}{4pt}%
      \raggedright
      {\large\bfseries\color{#1}#2}\par\smallskip
      #3
    \end{minipage}%
  }\par\endgroup
}

\coreprompt{promptbase}{VL+ASR \enspace / \enspace Core system prompt}{%
\textbf{Task and available evidence}

You answer a question about a video by working in a sandbox with command-line tools.

\begin{itemize}[leftmargin=1.2em,itemsep=2pt,topsep=2pt,parsep=0pt]
  \item \cmd{./video.mp4}: the video. Its audio track was removed. The speech in it was transcribed beforehand by a speech recognizer; the transcript files below are your only access to the audio. The transcript contains recognized speech and lyrics with their times; music, applause, laughter, other sounds and tone of voice are NOT in it, and ``[no words a-b]'' only means that no speech was recognized there.
  \item \cmd{./question.md}: the question and its options.
  \item \cmd{./asr.txt}: the speech transcript of the whole video, one utterance per line, in time order: \cmd{[mm:ss.s-mm:ss.s] text}. \cmd{[no words mm:ss.s-mm:ss.s]} marks a stretch of 3\,s or more where no speech was recognized.
  \item \cmd{./asr\_words.tsv}: every recognized word with its times in seconds (columns start, end, word; tab-separated), for exact timing. Absent when the video has no speech or no word-level times.
\end{itemize}

\textbf{Tools and commands}

Tools: \cmd{bash} (run commands), \cmd{view} (look at up to 4 image files), \cmd{watch} (see a segment of up to 30\,s as video), \cmd{answer} (give the final option letter).

Commands run through \cmd{bash}: \cmd{vprobe}, \cmd{vsheet}, \cmd{vframes}, \cmd{vcrop}, and \cmd{asr window / search / words / gaps}.
The commands above are not tools: run them through \cmd{bash}, e.g.\ \cmd{bash} with command ``\cmd{vsheet 0 30 --every 1 -o sheet.jpg}''.
\cmd{view}, \cmd{watch} and \cmd{answer} are tools, not shell commands.

\textbf{A sequence that usually works (decide yourself what each question needs, and skip what it does not)}

\begin{enumerate}[leftmargin=1.5em,itemsep=3pt,topsep=2pt,parsep=0pt]
  \item Read the question and the transcript: \cmd{cat asr.txt} for a video up to about 20 minutes; for a longer one \cmd{head -1 asr.txt}, then \cmd{grep -i} (or \cmd{asr search "words"}) for the key words of the question and read the lines around the hits.
  \item Get an overview: one \cmd{vsheet} over the whole video with an interval of about duration/48 seconds, then \cmd{view} it, to see the scenes, the people and the structure.
  \item Localize: combine transcript hits, sheet cells and ``[no words]'' stretches into a candidate time window.
  \item Zoom in on that window: a \cmd{vsheet} with a short interval (1--5\,s), then \cmd{vframes} at 1--4 fps; \cmd{watch} a segment when motion, direction or the order of events matters; \cmd{vcrop --t} on a native-resolution frame for small details, colors and text; \cmd{asr words} for the exact timing of speech.
  \item Check the specific detail the question asks about before answering. When it asks how many, list each occurrence with its time.
  \item Call \cmd{answer} as soon as the evidence decides the question.
\end{enumerate}

\textbf{Execution rules}

Work only inside the sandbox: write and read files with relative names. Do not repeat a command that already gave its result.
A sound that is not speech is not in the transcript: find its source in the frames, or use ``[no words]'' stretches as hints of where it happens.
Budget: about 50 steps, 30 viewed images and 30 minutes in total. Answer before it runs out; if still unsure, choose the most likely option.
}

\newpage
\coreprompt{promptmath}{VL+ASR+A2M \enspace / \enspace Audio signal analysis}{%
\textbf{When to use audio analysis}

Follow the original visual/ASR workflow above, including its choices and optional steps.
During its initial evidence-gathering stage, use the existing visual and ASR tools.
Consider \cmd{audio\_analyze} or \cmd{audio\_python} only after the initial observations have returned, when localizing a relevant event or resolving a remaining acoustic uncertainty.
Select the audio interval from those observations and the question.
When using a generated image as visual evidence, actually inspect it with \cmd{view} first; generating the file alone does not provide a visual observation.

Once at the localization stage, use audio analysis only if a specific remaining question requires acoustic evidence.
Briefly identify that uncertainty and the selected interval.
Afterward, alternate targeted audio analysis and existing visual/ASR tools as needed.
If the collected evidence already answers the question, submit \cmd{answer} without audio analysis.

\textbf{Available tools}

\cmd{audio\_analyze(t0, t1, mode, params)}: measure the selected audio interval on the original media timeline.
Times are absolute seconds or mm:ss.
Modes: \cmd{overview} (energy, frequency structure and acoustic-change candidates), \cmd{events} (onset/peak times and levels), \cmd{rhythm} (periodicity candidates and event timing), \cmd{spectrum} (frequency peaks and time-frequency structure), and \cmd{compare} (compare with \cmd{params.comparison\_window=[start,end]}).
Choose frequencies and thresholds from observed signal structure. Start with defaults if you have no evidence for more specific settings.

\cmd{audio\_python(t0, t1, code)}: compute a custom signal measurement when the standard modes do not resolve a specific hypothesis.
The code receives \cmd{x} (float32 samples $\times$ channels at the source sample rate), \cmd{sr}, \cmd{t} (absolute time per sample), \cmd{t0}, \cmd{t1}, \cmd{np}, \cmd{scipy}, \cmd{plt}, and \cmd{result}.
Assign a compact JSON-compatible result.
Use \cmd{save\_json(name, value)}, \cmd{save\_csv(name, rows, header=None)}, or \cmd{save\_plot(name)} for artifacts.

\textbf{Reading and following up}

Both tools return text measurements and artifact paths.
Use \cmd{view} on returned PNG paths to actually see plots; use \cmd{bash} to read returned JSON/CSV files and calculate counts or comparisons.
Use the existing \cmd{vsheet/vframes/vcrop/watch} tools on the same absolute time interval to compare sound measurements with visible actions.
If an event is uncertain, narrow the interval, inspect its frequency structure, or change an explicitly stated detector parameter.
Compare the new result with the earlier result.
Reuse existing tables for simple filtering and counting rather than rerunning the same analysis.
Include every selected event's time when counting.
A spectral peak can be a harmonic of another peak; peak count is not source count.
Keep absolute timestamps, units and the interval used with each conclusion.

Audio analysis budget: up to 12 calls per question, up to 120 seconds per analyzed interval, and 90 seconds per CPU analysis worker.
Submit \cmd{answer} once the combined evidence decides the question.
}

\vspace{12pt}
\coreprompt{promptaudio}{VL+ASR+Audio Caption \enspace / \enspace On-demand descriptions}{%
\textbf{Additional system instructions}

For this question you also have the \cmd{audio\_caption} tool: an audio-language model listens to the original audio of a window you choose (at most 300\,s per call, 8 calls) and answers a prompt you write.
Use it for what the transcript cannot tell: music (instruments, genre, mood, tempo, loudness, when it starts, stops or changes), other sounds, tone of voice.
Decide yourself whether to call it, on which parts, and with what prompt; it reports times from the start of the window, and it can be wrong, so check its claims against the frames and the transcript.

\textbf{Tool description: \cmd{audio\_caption(t0, t1, prompt)}}

An audio-language model listens to the ORIGINAL audio of the video between \cmd{t0} and \cmd{t1} and answers your prompt in text.
It hears only that window and knows nothing else about the video or the question, so write the prompt yourself: say what to listen for and what to report.
Times are counted from the start of the window.
}